\documentclass[a4paper,fleqn]{cas-dc}
\usepackage{caption}
\usepackage[authoryear,longnamesfirst]{natbib}
\usepackage{amsmath,amssymb,amsfonts}
\usepackage{graphicx}
\usepackage{subcaption}
\usepackage[table]{xcolor}
\usepackage{booktabs}
\usepackage{multirow}
\usepackage{array}
\usepackage{colortbl}
\usepackage{pifont}
\usepackage{adjustbox}
\usepackage{placeins}
\usepackage{makecell}
\usepackage{textcomp}
\usepackage{algorithmic}

\graphicspath{{./}{figures/}}

\definecolor{i1}{RGB}{245,245,245}
\definecolor{i2}{RGB}{232,241,255}
\newcommand{\sepvert}{\hspace{2pt}\vrule\hspace{2pt}}

\begin{document}
\let\WriteBookmarks\relax
\def\floatpagepagefraction{1}
\def\textpagefraction{.001}
\setcounter{topnumber}{5}
\setcounter{dbltopnumber}{5}
\renewcommand{\topfraction}{0.95}
\renewcommand{\dbltopfraction}{0.95}
\renewcommand{\textfraction}{0.05}
\renewcommand{\floatpagefraction}{0.90}
\renewcommand{\dblfloatpagefraction}{0.90}

\shorttitle{MedLVR: Latent Visual Reasoning for Medical Visual Question Answering}
\shortauthors{Xi et~al.}

\title[mode=title]{MedLVR: Latent Visual Reasoning for Medical Visual Question Answering}
\tnotemark[1]
\tnotetext[1]{This work was supported in part by the National Institutes of Health under award numbers R01CA272991, R01DE033512, and R01EB032680.}

\author[1]{Suyang Xi}
\fnmark[1]

\author[1]{Songtao Hu}
\fnmark[1]

\author[1]{Yuxiang Lai}

\author[2]{Wangyun Dan}

\author[2]{Yaqi Liu}

\author[1]{Shansong Wang}

\author[1]{Xiaofeng Yang}
\cormark[1]
\ead{xiaofeng.yang@emory.edu}

\affiliation[1]{
    organization={Department of Radiation Oncology and Winship Cancer Institute, Emory University School of Medicine},
    city={Atlanta},
    postcode={30322},
    state={GA},
    country={USA}
}

\affiliation[2]{
    organization={Department of Biostatistics and Bioinformatics, Emory University},
    city={Atlanta},
    postcode={30322},
    state={GA},
    country={USA}
}

\cortext[1]{Corresponding author.}
\fntext[1]{Suyang Xi and Songtao Hu contributed equally to this work.}

\begin{abstract}
Medical vision--language models (VLMs) have shown strong potential for medical visual question answering (VQA), yet their reasoning remains largely text-centric: images are encoded once as context, while subsequent inference is carried primarily by language. This can be limiting in clinical scenarios, where accurate answers often depend on subtle and localized visual cues whose relevance may only become clear as the question is interpreted. We propose \textsc{MedLVR}, a latent visual reasoning framework that introduces a continuous latent pathway into autoregressive decoding. Rather than relying only on textual intermediate reasoning, \textsc{MedLVR} inserts a short latent segment by reusing decoder hidden states as continuous reasoning steps, providing additional internal computation over query-relevant visual information before answer generation. We adopt a two-stage training strategy. Region-of-interest (ROI)-supervised fine-tuning shapes latent states toward question-associated visual evidence, while Visual--Latent Policy Optimization (VLPO) extends outcome-level optimization to the latent trajectory together with answer generation. Experiments on OmniMedVQA and five external medical VQA benchmarks show consistent gains over the Qwen2.5-VL-7B backbone, improving the average external-benchmark score from 48.3\% to 53.4\%. These results support latent visual reasoning as an effective way to strengthen the use of visual evidence during medical VQA.
\end{abstract}


\maketitle

\section{Introduction}
\label{sec:introduction}

\begin{figure*}[t]
    \centering
    \includegraphics[width=\textwidth]{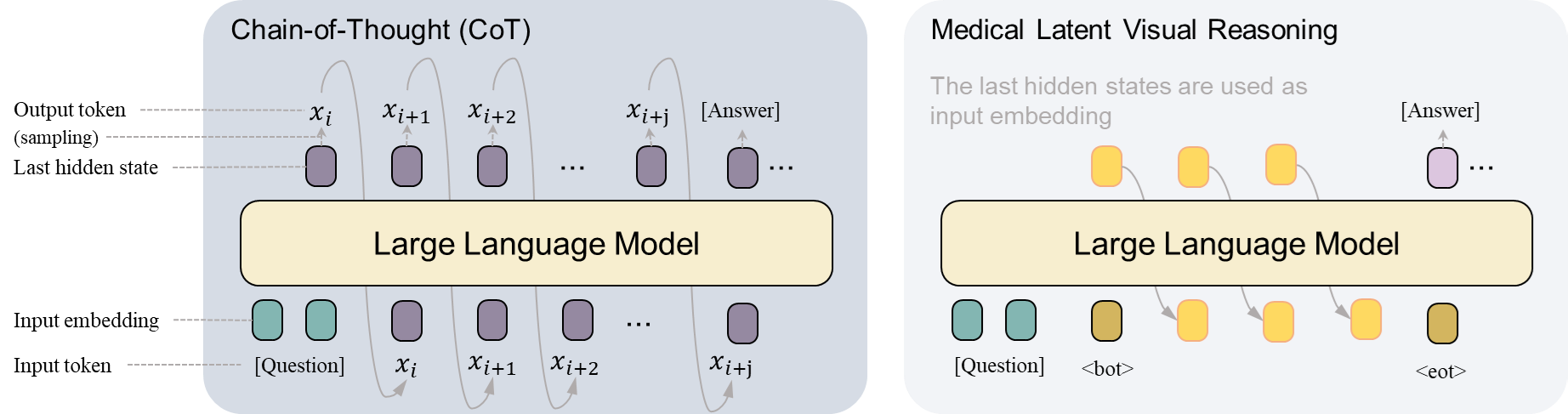}
    \caption{Comparison between conventional text-space reasoning and Medical Latent Visual Reasoning. In conventional text-space reasoning, the model autoregressively generates intermediate textual reasoning tokens before producing the final answer. In Medical Latent Visual Reasoning, the last hidden states are directly reused as input embeddings to form latent reasoning steps, providing additional hidden-space computation before answer generation.}
    \label{fig:mlvr_vs_cot}
\end{figure*}

Vision--language models (VLMs) have achieved impressive results on natural-image tasks, including visual question answering and multimodal dialogue \cite{radford2021learning,li2023blip,liu2023visual}. However, directly extending this paradigm to medical imaging remains challenging \cite{fu2024blink,tong2024eyes}. Medical visual question answering often depends on a small number of subtle and highly localized visual cues \cite{bigverdi2025perception}, whose diagnostic significance may not be fully determined when the image is first encoded. A local visual pattern may initially appear ambiguous and become clinically meaningful only after being interpreted in relation to the question and the surrounding image context. Medical visual reasoning therefore requires more than extracting visual features once; question-relevant information may need to remain available and be progressively refined as inference unfolds.

Most existing medical VLMs nevertheless encode the image once as visual context \cite{team2024chameleon,li2024llava} and subsequently rely primarily on autoregressive text generation for reasoning \cite{bai2025qwen2,wang2025internvl3,liu2023visual}. This creates a structural asymmetry: the linguistic state evolves throughout decoding, whereas visual evidence largely remains within the context established before reasoning begins. For medical imaging, the central issue is therefore not only whether the model has perceived the correct content at the outset, but whether that visual information retains an internal representation that can continue to develop during inference.

Recent methods that strengthen the interaction between vision and reasoning largely follow two paradigms \cite{zhang2023multimodal,shao2024visual,su2025openthinkimg}. The first is \textit{Thinking about Images}. These methods expand intermediate reasoning in language space through multimodal chains of thought, supervised reasoning traces, or reinforcement-learning-based post-training \cite{shao2024visual,xu2025llava}. A key limitation is that intermediate reasoning is still carried primarily by language. Visual evidence must first be converted into text-compatible semantics before it can participate in subsequent inference. For subtle medical patterns that are difficult to describe precisely, this conversion may force an early semantic commitment before the diagnostic significance of the evidence has been sufficiently resolved. In other words, although the reasoning chain becomes longer, what is expanded is primarily linguistic computation; visual information can continue to participate only after it has been verbalized \cite{tong2024eyes,huang2024opera}.

The second paradigm is \textit{Thinking with Images}. These methods reacquire local visual information through operations such as cropping, zooming, and region revisiting \cite{zhang2025chain,su2025openthinkimg,fu2025refocus}, or through more general tool-based interaction with the visual input \cite{suris2023vipergpt,chen2025think,lu2026medvistagym}. The resulting views are then fed back into the model for further reasoning, often yielding more explicit region-level visual grounding \cite{wang2025monet}. However, this grounding is achieved primarily through external visual operations rather than through the continued evolution of visual representations within the reasoning process. Each evidence update typically requires the model to plan a visual action, obtain a new view, re-encode the image, and resume reasoning. Thinking with Images therefore retains a repeated encode--reason cycle, in which new visual inputs are constructed and re-encoded through external operations. These repeated perception--generation loops introduce substantial time and computational overhead, while making evidence updates dependent on discrete external interactions rather than a continuous internal visual reasoning process.

Current medical VLMs are therefore largely constrained to two choices: reasoning \textit{about} images in language space or repeatedly reasoning \textit{with} images through external visual operations. The former extends reasoning by generating more language, whereas the latter extends perception by acquiring more views. The former requires visual evidence to be expressed through language, whereas the latter can explicitly access relevant regions but depends on additional visual operations and re-encoding. Neither directly targets continuous internal computation over question-relevant visual evidence. This leads to a natural question:

\begin{quote}
\textit{Can medical visual question answering allow question-relevant visual evidence to remain influential throughout internal reasoning, without requiring that intermediate evidence be fully verbalized or repeatedly reacquired through external visual operations?}
\end{quote}

Continuous latent representations provide a possible route toward this goal. Recent studies have shown that intermediate reasoning can be carried out in continuous hidden space \cite{hao2024training,cheng2024compressed,shen2025codi}. In multimodal settings, related work has further explored the use of latent states to carry visual information \cite{li2025latent,wang2025monet}. However, showing that latent space can support reasoning is not sufficient to address the central problem in medical VQA. Without explicit visual constraints, continuous states may degenerate into arbitrary hidden computation. Even when a latent trajectory exists, conventional objectives mainly supervise the final answer and do not directly shape the process that precedes it. The key challenge is therefore not simply to insert additional latent steps, but to give the latent trajectory an explicit question-relevant visual target and to optimize the task utility of the resulting internal process.

Motivated by this observation, we introduce \textbf{MedLVR}, a distinct reasoning pathway for medical VQA that we term \textit{Thinking in Latent Visual Space}. Unlike Thinking about Images, which unfolds intermediate reasoning as text, and Thinking with Images, which repeatedly reacquires visual information through external operations, MedLVR inserts a short sequence of continuous latent reasoning steps into autoregressive decoding. These states remain internal to the model and are progressively propagated before the final answer is generated, allowing question-relevant visual information to participate directly in intermediate inference. MedLVR therefore expands neither the length of textual reasoning nor the number of external perception steps. Instead, it expands the internal computation performed before the final answer is committed to language.

To shape this latent process toward question-associated visual evidence, we first introduce ROI-supervised fine-tuning. Here, the ROI is not merely a localization label that tells the model \textit{where to look}. More importantly, it defines an \textit{evidence formation target} for continuous latent reasoning: after sufficient latent-state evolution, the internal representation should increasingly reflect visual content associated with the current question. The ROI therefore does not prescribe an explicit visual action. Instead, it specifies what kind of visual evidence should emerge from a sufficiently developed latent reasoning process. In this sense, Thinking with Images obtains additional evidence through external visual operations, whereas MedLVR uses continuous latent computation to shape the internal representation of question-associated visual information.

The introduction of latent visual reasoning also changes the optimization problem. Conventional supervised learning and reinforcement learning primarily act on the final textual answer, leaving intermediate latent computation to receive only indirect supervision through the output. We therefore further employ \textbf{Visual-Latent Policy Optimization (VLPO)}, which extends outcome-level optimization to both the latent trajectory and subsequent answer generation. This allows task rewards to directly influence the continuous latent process that precedes the final answer \cite{wang2025monet,chen2025think}. The resulting two-stage strategy separates two roles: the first stage shapes what visual information is represented along the latent trajectory, while the second optimizes whether the resulting trajectory contributes to a successful answer.

In summary, our main contributions are three-fold:

\begin{enumerate}
    \item \textbf{A distinct reasoning pathway for medical VQA.}

    We introduce \textit{Thinking in Latent Visual Space} as a distinct reasoning pathway that complements \textit{Thinking about Images} and \textit{Thinking with Images}, allowing question-relevant visual evidence to remain influential during intermediate inference without being fully verbalized or repeatedly reacquired through external operations. Concretely, our ROI-conditioned evidence-formation objective (Eq.~\ref{eq:evidence_formation_loss}) uses region annotations as training-time visual references to guide the multimodal latent trajectory, without requiring ROI prediction or region inputs at inference.

     \item \textbf{Latent evidence formation through region-supervised reasoning.}
    We use ROI annotations as training-time visual references to guide the multimodal latent trajectory toward question-associated image evidence, encouraging the resulting latent states to retain information relevant to answer formation.
    \item \textbf{Joint optimization of latent reasoning and answer generation.}
    We further employ VLPO to extend outcome-level learning from final text generation to the latent trajectory, allowing the internal visual reasoning process to be optimized together with the answer.
    VLPO defines a Gaussian stochastic policy over continuous latent actions (Eq.~\ref{eq:latent_policy}) and uses matched-context trajectory replay (Eq.~\ref{eq:text_ratio}) to ensure that importance ratios isolate true policy changes from discrepancies caused by different preceding trajectories.
\end{enumerate}

\section{Related Work}

\noindent\textbf{Medical VLMs.} Early medical multimodal models typically adapt general-purpose VLMs to clinical settings through supervised fine-tuning and instruction alignment on medical corpora, such as LLaVA-Med \cite{li2023llava} and Med-Flamingo \cite{moor2023med}, aiming to better match medical QA and clinical language. More recently, models such as Med-Gemma \cite{sellergren2025medgemma} and Lingshu \cite{xu2025lingshu} place greater emphasis on medical-native pretraining or more systematic data design to broaden coverage of medical concepts and improve cross-task transferability. Meanwhile, the community has explored stronger alignment strategies, including reinforcement-learning post-training and more refined data construction \cite{yang2025r1,comanici2025gemini}. Overall, most approaches optimize final answers while placing only limited constraints on the intermediate use of visual evidence, leaving room for fluent but weakly grounded conclusions.

\noindent\textbf{Reasoning with Text and Tools.} \emph{Thinking about Images.} This line treats intermediate reasoning as a text-space training target and has evolved from SFT-based trace supervision to RL-based trajectory alignment. Multimodal CoT supervision and stepwise SFT strategies are commonly used to help models internalize explicit reasoning traces during training \cite{shao2024visual,xu2025llava}. More recent RL-style post-training further steers intermediate traces toward task-level objectives and robustness \cite{yang2025r1}. Beyond trace supervision, several works introduce explicit intermediate carriers such as visual scratchpads and annotations, so that the generated rationale is conditioned on actionable visual marks rather than purely linguistic scaffolds \cite{hu2024visual}. In parallel, hallucination-oriented objectives penalize over-reliance on language priors and encourage retrospection or improved attention allocation during generation \cite{huang2024opera}.

\emph{Thinking with Images.} This line makes multimodal reasoning \emph{procedural}: the model plans explicit perception actions, executes them, and re-conditions on the resulting visual states, rather than relying on a one-shot image embedding \cite{su2025openthinkimg}. A common instantiation is region revisiting, where the model proposes candidate ROIs, performs zooming or cropping, and feeds refreshed patches back to improve visual grounding \cite{zhang2025chain,su2025openthinkimg}. Tool-augmented variants extend the returned state beyond crops by invoking external executors or editors, producing intermediate artifacts such as code-generated outputs or edited views that are re-encoded as additional context \cite{suris2023vipergpt,fu2025refocus}. To standardize tool--model integration, recent work introduces structured primitives, including perception tokens and visualization-of-thought-style traces, so that intermediate visual states and their use are explicitly represented during reasoning \cite{bigverdi2025perception,li2025imagine}.

\noindent\textbf{Reasoning in Latent Space.} Recent work shifts intermediate reasoning from discrete tokens to continuous latent embeddings, treating it as a compact state trajectory learned via latent-space training and compression or self-distillation \cite{hao2024training,cheng2024compressed,shen2025codi}. This shift offers a different knob for scaling test-time compute: instead of lengthening textual chains, models allocate additional latent steps that refine internal states while keeping the external output concise \cite{geiping2025scaling}. In multimodal settings, latent visual reasoning uses these latent updates to refine internal representations of question-relevant visual information, interleaving latent refinement with standard decoding \cite{li2025latent,wang2025monet}. Building on this paradigm, we adapt latent reasoning to medical VQA and introduce objectives that shape latent trajectories toward localized visual evidence and directly optimize those trajectories for downstream task outcomes \cite{li2025latent,wang2025monet,yang2025r1}.

\begin{figure*}[tbp]
    \centering
    \includegraphics[width=\textwidth]{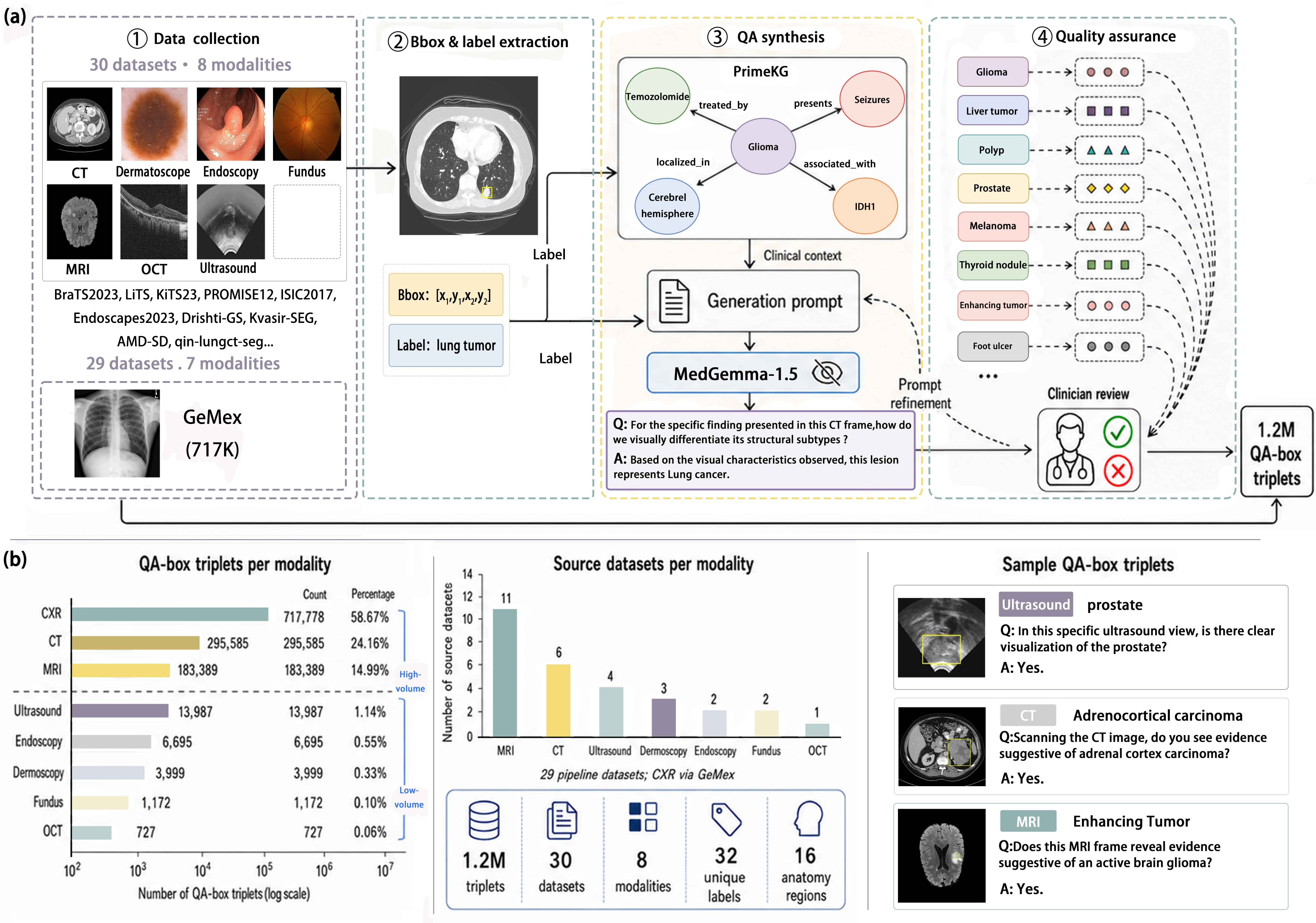}
    \caption{\textbf{Construction and composition of the large-scale medical grounding corpus.}
    \textbf{a}, Overview of the corpus construction pipeline. We collect 30 public
    medical imaging datasets spanning eight modalities. For the 29 non-CXR source
    datasets, expert spatial annotations are converted into bounding boxes and
    diagnostic labels. The labels are enriched with clinically relevant context
    retrieved from PrimeKG and passed to MedGemma-1.5 for text-only QA generation,
    while the raw image and bounding box coordinates are withheld from the
    generation model. The quality of the automatically generated QA–box triplets is estimated through a stratified expert audit of 1,200 samples. The GEMeX chest X-ray corpus is incorporated
    directly and unified with the automatically generated data, resulting in
    1,223,332 QA--box triplets.
    \textbf{b}, Composition and representative examples of the resulting corpus.
    The left panel shows the number and proportion of QA--box triplets across the
    eight imaging modalities; the middle panel summarizes the number of source
    datasets per modality and the overall corpus statistics; and the right panel
    presents representative triplets from ultrasound, CT, and MRI. In total, the
    corpus contains 505,554 automatically generated triplets from 29 non-CXR
    datasets together with 717,778 CXR pairs from GEMeX.}
    \label{fig:dataset_pipeline}
\end{figure*}

\section{Large-Scale Multi-Modal Medical Grounding Corpus}
\label{sec:dataset}

To support region-level visual supervision for latent reasoning, we construct a large-scale, fine-grained medical grounding corpus comprising approximately 1.2 million question--answer--box triplets spanning eight core clinical imaging modalities. This section details the data sources, the automated annotation pipeline, and the quality assurance procedure. An overview of the complete construction pipeline and the resulting modality distribution is provided in Fig.~\ref{fig:dataset_pipeline}.

\subsection{Data Sources and Modality Coverage}

Our corpus integrates 30 publicly available medical imaging datasets
across eight modalities: computed tomography (CT), chest X-ray (CXR),
dermoscopy, endoscopy, fundus photography, magnetic resonance imaging
(MRI), optical coherence tomography (OCT), and ultrasound. These
datasets were selected to maximize anatomical and pathological
diversity while ensuring that each source provides expert-verified
spatial annotations in the form of bounding boxes with associated
diagnostic labels.

For seven of the eight modalities (CT, dermoscopy, endoscopy, fundus
photography, MRI, OCT, and ultrasound), we apply our automated QA
generation pipeline described in Section~\ref{subsec:pipeline}. For
chest X-ray, we directly incorporate the GEMeX
dataset~\cite{11444163}, which provides high-quality, expert-curated
question--answer pairs grounded in radiographic findings. This design
choice reflects both the scale and maturity of existing CXR QA
resources, avoiding redundant generation where clinically validated
data already exists. As illustrated in Fig.~\ref{fig:dataset_pipeline},
the resulting corpus exhibits a modality distribution that reflects
the natural variation in publicly available segmentation and detection
datasets: CXR and CT together account for over 82\% of the total
volume (58.67\% and 24.14\%, respectively), MRI contributes 14.96\%,
and ultrasound accounts for 1.14\%, while less commonly annotated
modalities such as endoscopy (0.55\%), dermoscopy (0.33\%), fundus
photography (0.10\%), and OCT (0.06\%) contribute smaller but
clinically important subsets.

This distribution is intentionally left unbalanced to reflect
the natural availability of publicly annotated medical imaging
data rather than artificially resampled to achieve uniform
modality coverage. CXR and CT dominate the corpus because
they constitute the majority of publicly available datasets
with expert-verified spatial annotations. Forcing uniform
modality representation would require either aggressive
oversampling of low-frequency modalities, which risks
overfitting to a small set of repeated examples, or
undersampling of well-annotated modalities, which discards
valuable training signal. We instead preserve the natural
data distribution during Stage~1 training and rely on
Stage~2 reinforcement learning on OmniMedVQA, which provides
balanced modality coverage across all eight imaging domains,
to calibrate cross-modality performance. The empirical effect
of this design on low-frequency modalities is examined in
Section~\ref{sec:experiments}.

\subsection{Automated QA Generation Pipeline}
\label{subsec:pipeline}

We design an automated pipeline that synthesizes clinically grounded question--answer pairs from three complementary sources of information: (1) expert-annotated bounding boxes providing precise spatial localization, (2) diagnostic labels identifying the clinical finding within each annotated region, and (3) structured biomedical knowledge retrieved from a large-scale knowledge graph. As shown in the left portion of Fig.~\ref{fig:dataset_pipeline}, the pipeline proceeds through three sequential stages. A central design principle is that the language model never receives the raw image as input; instead, it operates exclusively on the textual label and the retrieved clinical context. This text-only generation strategy is intended to anchor generated QA pairs to label- and knowledge-based clinical content rather than to model-generated visual descriptions.

\subsubsection{Stage 1: Bounding Box and Label Extraction}

For each of the 29 non-CXR source datasets, we directly extract spatial
bounding box annotations and their associated diagnostic labels from the
original expert-curated ground truth. Segmentation masks, where available, are
converted into tight bounding boxes by computing the minimum enclosing
rectangle of each connected component. Each annotation is represented as a
structured tuple $(I,b,l)$, where $I$ denotes the source image,
$b=[x_1,y_1,x_2,y_2]$ specifies the bounding box coordinates in pixel space,
and $l$ denotes the corresponding region label, which may refer to
a pathological finding (\emph{e.g.}, ``glioma,'' ``liver tumor,''
``polyp'') or an anatomical structure of interest (\emph{e.g.},
``prostate,'' ``optic disc''). We use the term \emph{region label}
throughout to encompass both categories.

This procedure preserves the clinical fidelity of the original annotations:
no model-generated localization is introduced, and bounding-box precision is
determined solely by the quality of the source dataset annotations.

\begin{table*}[!ht]
\centering
\caption{Summary of the 29 source datasets (excluding GEMeX for CXR) used to construct the medical grounding corpus. Each dataset name is hyperlinked to its public source. We report the imaging modality, number of generated QA--box triplets, and annotated diagnostic categories. The grand total includes 717{,}778 GEMeX CXR pairs incorporated directly without automated QA generation (Section~\ref{subsec:gemex}).}
\label{tab:dataset_summary}
\vspace{4pt}
\renewcommand{\arraystretch}{1.12}
\footnotesize
\setlength{\tabcolsep}{5pt}
\begin{tabular}{@{} p{4.8cm} l r p{6.2cm} @{}}
\toprule
\textbf{Dataset} & \textbf{Modality} & \textbf{\# QA-Box} & \textbf{Diagnostic Categories} \\
\midrule

\multicolumn{4}{@{}l}{\cellcolor{gray!8}\textit{\textbf{Computed Tomography (CT)}}} \\[1pt]
\href{https://competitions.codalab.org/competitions/17094}{LiTS} & CT & 120{,}321 & Liver; Liver tumor \\
\href{http://medicaldecathlon.com/}{MSD-Task7} & CT & 99{,}708 & Pancreas; Pancreas tumor \\
\href{http://medicaldecathlon.com/}{MSD-Task8} & CT & 49{,}727 & Hepatic tumor \\
\href{http://medicaldecathlon.com/}{MSD-Task10} & CT & 17{,}257 & Colon cancer \\
\href{http://medicaldecathlon.com/}{MSD-Task6} & CT & 7{,}477 & Lung tumours \\
\href{https://www.cancerimagingarchive.net/collection/adrenal-acc-ki67-seg/}{qin-lungct-seg} & CT & 1{,}095 & Lung cancer \\
\midrule

\multicolumn{4}{@{}l}{\cellcolor{gray!8}\textit{\textbf{Magnetic Resonance Imaging (MRI)}}} \\[1pt]
\href{https://www.synapse.org/brats2023}{BraTS2023-GLI} & MRI & 51{,}017 & Enhancing tumor \\
\href{https://www.synapse.org/brats2023}{BraTS2023-MEN} & MRI & 63{,}878 & Enhancing tumor \\
\href{https://www.synapse.org/brats2023}{BraTS2023-PED} & MRI & 5{,}566 & Enhancing tumor \\
\href{https://www.synapse.org/Synapse:syn51186045/wiki/621356}{MVSeg2023} & MRI & 34{,}984 & Posterior leaflet; Anterior leaflet \\
\href{https://www.cancerimagingarchive.net/collection/adrenal-acc-ki67-seg/}{Adrenal-ACC-Ki67-Seg} & MRI & 14{,}012 & Adrenocortical carcinoma \\
\href{https://muregpro.github.io/data.html}{$\mu$-RegPro-T2} & MRI & 4{,}220 & Prostate \\
\href{https://qubiq21.grand-challenge.org/}{QUBIQ2021} & MRI & 3{,}994 & Pancreatic lesion \\
\href{https://www.cancerimagingarchive.net/analysis-result/ispy1-tumor-seg-radiomics/}{ISPY1-Tumor-SEG} & MRI & 3{,}147 & Breast tumor \\
\href{https://promise12.grand-challenge.org/}{PROMISE12} & MRI & 1{,}473 & Prostate \\
\href{https://www.kaggle.com/datasets/haithem1999/prostate-annotated-dataset-for-image-segmentation?resource=download}{Prostate-Seg} & MRI & 950 & Prostate \\
\href{https://promise09.grand-challenge.org/}{PROMISE09} & MRI & 148 & Prostate \\
\midrule

\multicolumn{4}{@{}l}{\cellcolor{gray!8}\textit{\textbf{Ultrasound (US)}}} \\[1pt]
\href{https://muregpro.github.io/data.html}{$\mu$-RegPro-US} & US & 4{,}586 & Prostate \\
\href{https://github.com/haifangong/TRFE-Net-for-thyroid-nodule-segmentation}{TG3K} & US & 3{,}585 & Thyroid gland \\
\href{https://github.com/haifangong/TRFE-Net-for-thyroid-nodule-segmentation}{TN3K} & US & 3{,}493 & Thyroid nodule \\
\href{https://www.kaggle.com/c/ultrasound-nerve-segmentation}{Nerve Segmentation} & US & 2{,}323 & Brachial plexus \\
\midrule

\multicolumn{4}{@{}l}{\cellcolor{gray!8}\textit{\textbf{Dermoscopy (Derm)}}} \\[1pt]
\href{https://challenge.isic-archive.com/landing/2017/}{ISIC2017} & Derm & 2{,}750 & Benign nevi; Melanoma; Seborrheic keratosis \\
\href{https://fusc.grand-challenge.org/}{Foot Ulcer Seg} & Derm & 1{,}138 & Foot ulcer \\
\href{https://www.fc.up.pt/addi/ph2\%20database.html}{PH$^{2}$} & Derm & 111 & Common nevus \\
\midrule

\multicolumn{4}{@{}l}{\cellcolor{gray!8}\textit{\textbf{Endoscopy (Endo)}}} \\[1pt]
\href{https://github.com/CAMMA-public/Endoscapes?tab=readme-ov-file}{Endoscapes2023} & Endo & 5{,}695 & Calot's triangle; Cystic artery; Cystic duct; Gallbladder; Cystic plate \\
\href{https://datasets.simula.no/kvasir-seg/}{Kvasir-SEG} & Endo & 1{,}000 & Polyp \\
\midrule

\multicolumn{4}{@{}l}{\cellcolor{gray!8}\textit{\textbf{Fundus Photography}}} \\[1pt]
\href{https://github.com/miag-ull/rim-one-dl}{RIM-ONE DL} & Fundus & 970 & Optic cup; Optic disc \\
\href{https://cvit.iiit.ac.in/projects/mip/drishti-gs/}{Drishti-GS} & Fundus & 202 & Optic cup; Optic disc \\
\midrule

\multicolumn{4}{@{}l}{\cellcolor{gray!8}\textit{\textbf{Optical Coherence Tomography (OCT)}}} \\[1pt]
\href{https://springernature.figshare.com/articles/dataset/An_Optical_Coherence_Tomography_Image_Dataset_for_wet_AMD_Lesions_Segmentation/25513435?backTo=%2Fcollections%2FAMD-SD_An_Optical_Coherence_Tomography_Image_Dataset_for_wet_AMD_Lesions_Segmentation%2F7157554&file=48777037}{AMD-SD} & OCT & 727 & Pigment epithelial detachment \\

\midrule\midrule
\multicolumn{2}{@{}l}{\textbf{Subtotal (29 datasets)}} & \textbf{505{,}554} & \\
\multicolumn{2}{@{}l}{\textbf{+ \href{https://huggingface.co/datasets/BoKelvin/GEMeX-VQA}{GEMeX} (CXR)}} & \textbf{717{,}778} & \\
\multicolumn{2}{@{}l}{\textbf{Grand Total}} & \textbf{1{,}223{,}332} & \\
\bottomrule
\end{tabular}
\end{table*}

\subsubsection{Stage 2: Knowledge Graph Enrichment via PrimeKG}

To augment each diagnostic label with structured clinical context, we query PrimeKG~\cite{chandak2022building}, a precision medicine knowledge graph that integrates 20 high-quality biomedical resources and encodes over 100{,}000 nodes across ten major biological entity types, connected by approximately 8 million edges representing 30 relation types.

Given a diagnostic label $l$, we perform entity linking by matching $l$ against the disease and phenotype nodes in PrimeKG using exact string matching followed by fuzzy matching with a Levenshtein similarity threshold of 0.85 to accommodate nomenclature variations (\emph{e.g.}, ``hepatocellular carcinoma'' vs.\ ``HCC''). For each successfully linked entity $e_l$, we retrieve its local neighborhood within a two-hop radius, extracting relational triplets of the form $(e_l, r, e_t)$, where $r$ denotes the relation type and $e_t$ denotes the target entity. The retrieved relations span multiple clinically relevant categories, including:

\begin{itemize}
    \item \textbf{Disease--Drug}: therapeutic associations (\emph{e.g.}, glioma $\xrightarrow{\text{treated\_by}}$ temozolomide);
    \item \textbf{Disease--Symptom}: clinical manifestations (\emph{e.g.}, glioma $\xrightarrow{\text{presents}}$ seizures);
    \item \textbf{Disease--Anatomy}: anatomical localization (\emph{e.g.}, glioma $\xrightarrow{\text{localized\_in}}$ cerebral hemisphere);
    \item \textbf{Disease--Gene/Protein}: molecular associations (\emph{e.g.}, glioma $\xrightarrow{\text{associated\_with}}$ IDH1);
    \item \textbf{Drug--Side Effect}: adverse reaction profiles (\emph{e.g.}, temozolomide $\xrightarrow{\text{causes}}$ myelosuppression).
\end{itemize}

The retrieved triplets are serialized into a structured textual
context $\mathcal{K}_l$ that accompanies the diagnostic label in
the subsequent QA generation stage. To prevent knowledge overload,
we apply a relevance-based filtering step that retains at most 15
triplets per entity, prioritizing direct (one-hop) relations and
relations with higher edge confidence scores in PrimeKG.
Importantly, the knowledge context serves to inform the language
model about the clinical setting of each finding so that the
generated questions reflect appropriate medical terminology and
diagnostic reasoning. It does not dictate the content of the
generated questions: the prompt explicitly constrains question
generation to visually observable characteristics of the finding
(Section~\ref{subsec:pipeline}, Stage~3), and relations that describe
non-visual properties such as molecular associations or drug
side effects are used only to provide background context rather
than to generate questions about those properties directly.
Questions that test non-visual medical knowledge without requiring
image inspection are identified and excluded through the visual
reasoning necessity criterion in the expert audit
(Section~\ref{subsec:qa}).

\subsubsection{Stage 3: Text-Only QA Generation via MedGemma}

We employ MedGemma-1.5~\cite{sellergren2026medgemma} as the backbone language
model for QA pair generation. MedGemma receives only the diagnostic
label $l$, the imaging modality, and the retrieved knowledge context
$\mathcal{K}_l$ as input, while both the raw image $I$ and the bounding box
coordinates $b$ are withheld. By restricting the generation model to text-only inputs, we reduce
dependence on automatically generated image descriptions, which may
introduce perceptual hallucinations. However, text-only generation
does not by itself guarantee instance-level visual consistency.
Because MedGemma receives only the diagnostic label and clinical
context without observing the specific image, the generated questions
predominantly reflect \emph{label-level} visual attributes that are
typical of a given diagnostic category (e.g., ``What is the shape of
this lesion?'') rather than \emph{instance-level} attributes tied to
a specific image (e.g., ``Is this lesion larger than the adjacent
structure?''). Questions about category-typical appearance,
morphology, and tissue relationships are generally appropriate for
the annotated finding, whereas questions requiring precise
instance-specific information such as exact size, laterality, or
spatial extent relative to neighboring anatomy may not always match
the particular image.

For each annotated region $(I, b, l)$, we construct a structured
prompt that provides the modality, diagnostic label, and the
serialized clinical knowledge triplets from $\mathcal{K}_l$. The
prompt instructs the model to generate a clinically grounded
question--answer pair about the annotated finding. The specific
prompt formulation varies across diagnostic labels and modalities
to accommodate differences in clinical context and question
complexity.

An important design consideration is that text-only generation
can produce questions answerable from medical knowledge alone,
without requiring inspection of the image. We address this risk
through three complementary mechanisms. First, the prompt instructs the model to generate questions oriented toward
observable visual characteristics of the finding rather than questions
about general pathophysiology, treatment, or molecular associations
that could be answered from a textbook. In practice, this encourages
questions about category-typical visual attributes such as general
appearance, common morphological features, and expected anatomical
localization. We note that this design involves an inherent trade-off:
text-only generation avoids perceptual hallucinations from unreliable
image captioning, but the resulting questions are anchored to what is
generally true of the diagnostic category rather than what is
specifically visible in each individual image. Instance-specific
attributes such as precise lesion dimensions, laterality, or spatial
extent relative to neighboring structures could be more reliably
derived from structured computation over segmentation masks or
bounding box coordinates, a direction we leave for future work. Second, because each QA pair is coupled with a
bounding box that is used only as a training supervision signal,
the model cannot exploit the label at inference time: it
receives only the image and question, and must localize the
relevant region autonomously. A question that is trivially
answerable from the label during generation therefore still
requires visual grounding during training and inference, since
the label itself is never provided as input to the model.
Third, we include visual reasoning necessity as an explicit
criterion in our expert quality audit
(Section~\ref{subsec:qa}), where medical experts verify that
each sampled question requires inspecting the corresponding
image region to be answered correctly. Among the 1{,}200
audited examples, 93.4\% pass this criterion, confirming that
the large majority of the generated corpus requires visual
evidence beyond what the label and knowledge context alone can
provide.

The bounding box annotation $b$ is never exposed to the generation
model and is used exclusively as a training supervision signal
during ROI-supervised fine-tuning
(Section~\ref{sec:latent_evidence_formation}). At inference time, neither the
bounding box nor the diagnostic label is provided. The model must therefore infer the relevant visual evidence from the
image itself when answering the question.
This implicit grounding design stands in contrast to approaches
that explicitly prompt the model with spatial coordinates or
region markers, which risk teaching the model to rely on
externally provided cues rather than developing its own capacity
for visual localization.

For each annotated region, we generate one QA pair using nucleus
sampling with $p = 0.95$ and a temperature of $0.7$.
Post-generation, we apply rule-based deduplication using
token-level Jaccard similarity (threshold $> 0.8$) to remove
near-duplicate pairs across regions that share the same
diagnostic label.

\subsection{Integration of the GEMeX Chest X-Ray Corpus}
\label{subsec:gemex}

For the CXR modality, we incorporate the GEMeX
dataset~\cite{11444163}, a large-scale chest X-ray VQA
benchmark built upon the Chest ImaGenome~\cite{wu2021chest}
through radiologist-guided region re-grounding and
GPT-4o-based question generation. GEMeX provides
717{,}778 question--answer pairs covering 151{,}025
images across four question types (open-ended, closed-ended,
single-choice, and multiple-choice), with each pair accompanied
by textual reasoning and region-level bounding box annotations.
Because GEMeX already provides expert-validated spatial
grounding, we directly integrate its QA pairs into our corpus
without applying our automated generation pipeline. We align
the GEMeX entries to our unified data format by mapping its
30 predefined anatomical regions to bounding box coordinates,
ensuring consistency with the annotation schema used for the
remaining seven modalities.




\subsection{Quality Assurance}
\label{subsec:qa}

To assess the quality of the final automatically generated corpus,
we conduct a stratified expert audit of 1{,}200 QA--box triplets
sampled from the 505{,}554 examples produced by our generation
pipeline. Sampling is stratified by imaging modality and diagnostic
category to ensure that every modality is represented in proportion
to its clinical importance rather than its corpus frequency. For
the four low-frequency modalities (dermoscopy, endoscopy, fundus
photography, and OCT), we over-sample to ensure a minimum of 100
audited examples per modality, so that quality estimates for these
subsets are not dominated by sampling noise. At a corpus-level
acceptance rate of 91.7\%, 1{,}200 samples yield a 95\% confidence
interval of $\pm$1.6 percentage points, providing a statistically
reliable estimate of overall corpus quality.

Two independent medical experts review each sampled example along four
dimensions. \emph{Factual correctness} evaluates whether the answer is
clinically consistent with the diagnostic label and retrieved knowledge
context. \emph{Visual grounding consistency} evaluates whether the question
concerns medical content supported by the annotated region.
\emph{Linguistic fluency} evaluates whether the QA pair is well formed and
unambiguous. \emph{Visual reasoning necessity} evaluates whether correctly
answering the question requires inspecting the corresponding image region,
as opposed to being answerable from general medical knowledge or the
question text alone.

Each criterion is scored as pass or fail, and an example is accepted only when
all four criteria are satisfied. We report the overall acceptance rate together
with the pass rate for each criterion. Inter-reader agreement is quantified
using Cohen's $\kappa$ based on the independent ratings, and disagreements are
resolved by consensus only after the initial assessment has been completed.

Among the 1,200 reviewed examples, \textbf{91.7}\% satisfy all four quality
criteria. The criterion-specific pass rates are \textbf{96.8}\% for factual
correctness, \textbf{95.6}\% for visual grounding consistency,
\textbf{98.1}\% for linguistic fluency, and \textbf{93.4}\% for visual
reasoning necessity. Inter-reader agreement reaches
Cohen's $\kappa=\textbf{0.85}$. The most common failure modes involve
knowledge graph mismatches, where the retrieved PrimeKG triplets
introduce clinical associations that do not apply to the specific
imaging context of the annotated region, and text-answerable questions,
where the generated question can be resolved from the diagnostic label
or general medical knowledge without requiring visual evidence from
the annotated region. Acceptance rates are consistent across modalities,
ranging from 89.2\% (OCT) to 93.1\% (CT), with no modality falling
below 88\%. This uniformity suggests that the generation pipeline
maintains stable quality across imaging domains despite substantial
differences in corpus size and diagnostic complexity.

Beyond the expert audit, we conduct two empirical diagnostics to
verify that the generated QA pairs require visual evidence at the
corpus level. We sample a held-out subset of 10{,}000 QA pairs
from the automatically generated examples, stratified by modality.
In the \emph{question-only} condition, the backbone model
(Qwen2.5-VL-7B) receives only the question without any image,
measuring the extent to which correct answers can be inferred from
question text or answer priors alone. In the \emph{image-shuffled}
condition, each question is paired with a randomly selected image
from a different sample within the same modality, controlling for
the possibility that questions are answerable from generic
modality-level visual features rather than instance-specific content.

The question-only condition achieves an accuracy of 28.4\%, compared
to 60.3\% when the matched image is provided, yielding an
image-dependency gap of 31.9 percentage points. The image-shuffled
condition achieves 33.7\%, substantially below the matched-image
accuracy. Together with the expert audit, these results indicate
that the majority of the automatically generated QA pairs cannot be
resolved from the question text alone or from an arbitrary image of
the same modality.


\subsection{Training--Evaluation Data Overlap Analysis}

Because our grounding corpus draws from publicly available
segmentation and detection datasets, potential image-level overlap
with downstream evaluation benchmarks must be carefully considered.
We analyze overlap risk for each evaluation benchmark as follows.

\textbf{Low-risk benchmarks.} VQA-RAD sources its images from
MedPix, a radiology teaching case database with no overlap with our
30 source datasets. PMC-VQA draws figures from PubMed Central
articles, MMMU (Health \& Medicine) uses examination-style images,
and MedXpertQA sources from medical licensing examinations. None
of these share image provenance with our training corpus.

\textbf{Benchmarks requiring closer examination.} SLAKE derives
its CT and MRI images from the Medical Segmentation Decathlon
(MSD), which also serves as the source for four tasks in our
grounding corpus (MSD-Task6, 7, 8, and 10). However, SLAKE
samples 2D axial slices from MSD 3D volumes and uses an
independent set of physician-authored VQA annotations that are
entirely distinct from our bounding-box-based QA pairs. Crucially,
our corpus uses MSD tasks covering lung, pancreas, hepatic, and
colon anatomy, whereas SLAKE primarily samples head, neck, and
abdominal slices. The anatomical and task-level overlap is therefore
limited, though we cannot rule out that individual slices may appear
in both sets.

For OmniMedVQA, which aggregates 73 classification datasets,
partial source-level overlap may exist with ISIC (dermoscopy),
Kvasir-SEG (endoscopy), and PH$^2$ (dermoscopy). We note that
these three modalities collectively account for less than 1\% of our
training corpus (dermoscopy 0.33\%, endoscopy 0.55\%), limiting the potential contribution of these sources to the overall Stage~1
training distribution.

\textbf{Structural safeguards.} Several design choices further
mitigate contamination risk. First, our QA generation pipeline
operates on bounding-box annotations and text-only clinical context,
producing grounding-oriented question--answer pairs that differ
fundamentally in format, content, and difficulty from the
classification-style questions used in OmniMedVQA and the
physician-authored questions in SLAKE. Second, our Stage~2
reinforcement learning uses only the open-access split of
OmniMedVQA with a standard 80/20 train--test partition, following
the established protocol in prior work. Third, MedLVR's
improvements are consistent across all five external benchmarks,
including the low-risk ones (VQA-RAD $+6.0$, PMC-VQA $+4.6$,
MMMU $+10.2$), which share no data provenance with our training
corpus. This cross-benchmark consistency supports the interpretation that the observed
gains extend beyond datasets with identified source-level overlap.

\textbf{Quantitative evidence from overlap-free benchmarks.}
To further assess whether source-level overlap contributes to
the reported improvements, we compare MedLVR's gains on
benchmarks with potential overlap against those with no overlap.
Among the five external benchmarks, VQA-RAD, PMC-VQA, MMMU
(Health \& Medicine), and MedXpertQA share no image provenance
with our training corpus. On these four overlap-free benchmarks,
MedLVR improves over Qwen2.5-VL-7B by $+6.0$, $+4.6$, $+10.2$,
and $+1.8$ points, respectively, yielding an average gain of
$+5.65$ points. On SLAKE, where partial source-level overlap
with MSD exists, the gain is $+2.7$ points. The improvement on the potentially overlapping benchmark is
\emph{smaller} than the average improvement on the four benchmarks with no
identified source-level overlap. This pattern does not suggest a disproportionate
benefit on the benchmark with potential overlap, although it cannot by itself
exclude image-level duplication.

For OmniMedVQA, the in-domain evaluation follows a standard
80/20 train--test partition applied to the open-access split,
which is the same protocol adopted by all compared baselines
including Med-R1 and ViTAR. The potentially overlapping
modalities (dermoscopy and endoscopy) account for less than
0.9\% of our Stage~1 training corpus and represent a small
fraction of the OmniMedVQA test set. We further note that
MedLVR's strongest modality-level gains on OmniMedVQA occur
on CT ($+20.0$ over backbone) and MRI ($+23.6$), neither of
which involves any overlap with the evaluation data, while
performance on dermoscopy and endoscopy-related modalities
does not show anomalous improvements relative to the overall
trend.

\subsection{License and Ethical Compliance}

All 30 source datasets used in our grounding corpus are publicly
available and were obtained in accordance with their respective
data use agreements. The majority are released under permissive
open-access licenses: the Medical Segmentation Decathlon tasks
are released under CC-BY-SA~4.0~\cite{antonelli2022medical}; ISIC2017 under
CC0~1.0; BraTS2023 under a Synapse data-use agreement
permitting non-commercial research; and LiTS under
CC-BY-NC-SA. The remaining datasets, including Kvasir-SEG,
PH$^2$, Endoscapes2023, Drishti-GS, RIM-ONE~DL, and others,
are publicly released for research and educational purposes.
GEMeX is distributed through PhysioNet under its standard
data use agreement. All usage in this work is strictly for
non-commercial academic research, consistent with the terms of
each source license. A complete per-dataset license summary is
provided in the supplementary material.

\begin{figure*}[t]
    \centering
    \includegraphics[width=\textwidth]{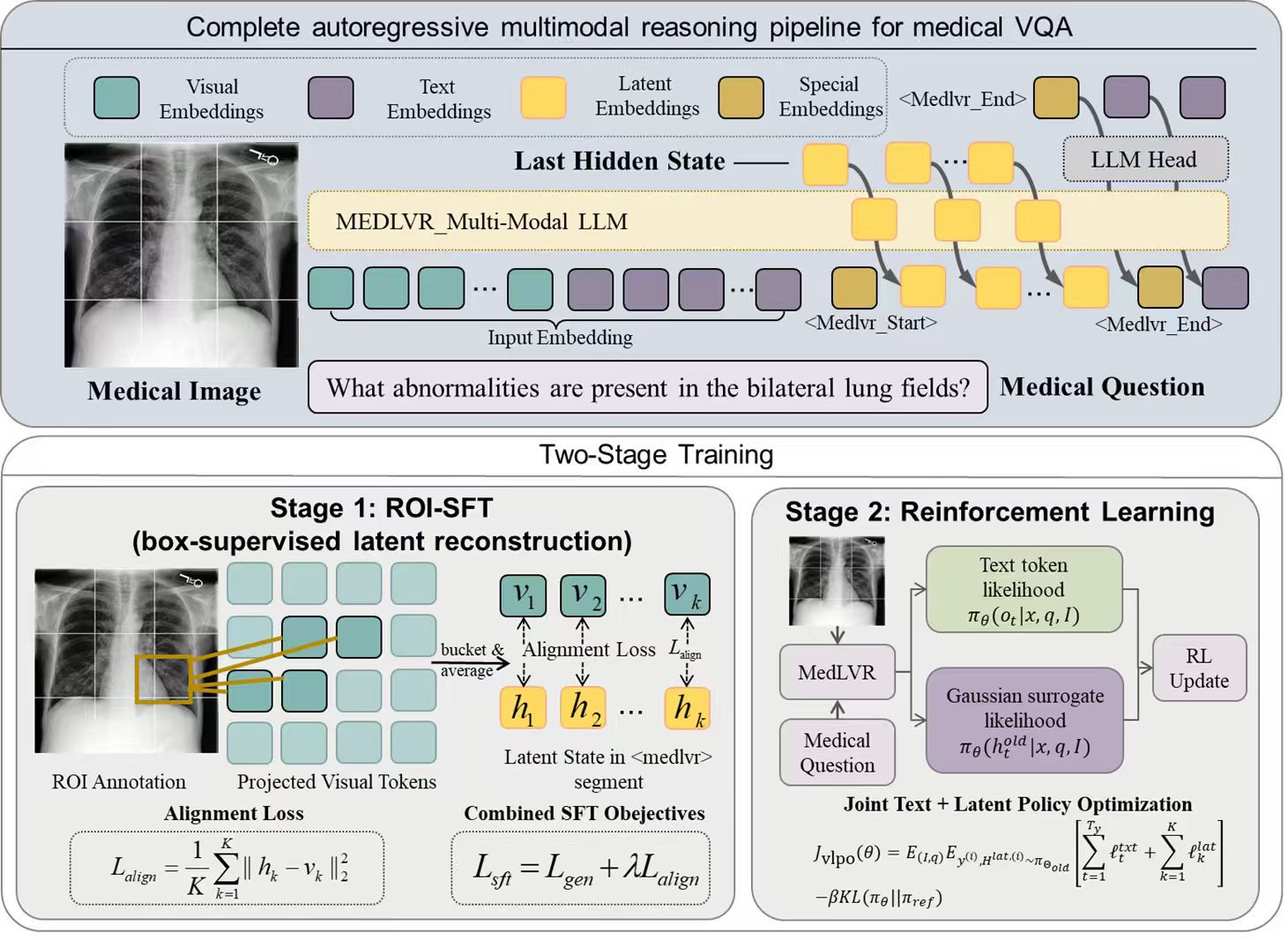}
    \caption{Overview of the proposed MedLVR framework for medical VQA. The top panel shows the autoregressive multimodal reasoning pipeline, in which visual, textual, latent, and special embeddings are jointly processed by the multimodal language model. Rather than generating long textual reasoning chains, MedLVR reuses hidden states between \texttt{<MedLVR\_Start>} and \texttt{<MedLVR\_End>} as latent reasoning steps before producing the final answer. The bottom panel illustrates the two-stage training strategy: Stage~1 uses ROI-supervised latent evidence formation to shape intermediate latent states, while Stage~2 applies reinforcement learning with the VLPO objective to directly optimize latent reasoning and answer generation.}
    \label{fig:MedLVR_framework}
\end{figure*}

\section{Method}
\label{sec:method}

\subsection{Medical Latent Visual Reasoning}
\label{sec:medical_lvr}

Fig.~\ref{fig:MedLVR_framework} presents the overall framework of
\textsc{MedLVR}. Given a medical image $I$ and a question $q$, the vision
encoder maps $I$ into $N$ patch-level visual tokens
$V=\{v_j\}_{j=1}^{N}$, where $v_j\in\mathbb{R}^{d_v}$ and $d_v$ denotes the
output dimension of the vision encoder. The backbone's native multimodal
projector $\mathrm{proj}(\cdot)$ maps these tokens into the language-model
space as $\tilde V=\mathrm{proj}(V)=\{\tilde v_j\}_{j=1}^{N}$, where
$\tilde v_j\in\mathbb{R}^{d}$ and $d$ is the decoder hidden dimension. Let
$T=\{t_m\}_{m=1}^{M}$ denote the embedded question tokens, with $M$ being
the number of textual input tokens. The projected visual tokens and question
tokens are concatenated into a unified multimodal context
$X_0=[\tilde V;T]$, which conditions all subsequent decoding steps.

Conventional medical VLMs largely perform reasoning after this visual context
has already been established. Increasing reasoning depth therefore typically
means generating additional linguistic steps. Methods that revisit the image
take a different route by acquiring new visual observations, but each update
requires another view to be constructed and encoded before reasoning can
continue. In both cases, the amount of internal computation performed directly
over visual evidence remains limited.

\textsc{MedLVR} introduces an alternative computation pathway. During
standard autoregressive decoding, once the model enters the latent reasoning
segment, ordinary vocabulary generation is temporarily suspended and the
decoder performs a fixed number of continuous latent updates. Let
$h_k\in\mathbb{R}^{d}$ denote the last-layer hidden state at the $k$-th latent
step. Rather than passing $h_k$ through the language-model head to generate a
vocabulary token, we directly reuse it as the input embedding of the next
latent step. The latent transition is written as
\begin{equation}
h_{k+1}
=
f_{\theta}
\!\left(
h_k;
X_0,
h_{<k}
\right),
\qquad
k=1,\ldots,K-1,
\label{eq:latent_transition}
\end{equation}
where $f_{\theta}$ denotes the multimodal decoder parameterized by $\theta$,
$h_{<k}=\{h_1,\ldots,h_{k-1}\}$ is the preceding latent prefix, and $K$ is
the fixed latent reasoning budget. The resulting deterministic trajectory used
in Stage~1 and at inference is
$H^{\mathrm{lat}}=\{h_k\}_{k=1}^{K}$.

After the $K$ latent updates are completed, decoding returns to the textual
stream and the model generates the answer
$y=\{y_t\}_{t=1}^{T_y}$, where $T_y$ denotes the answer length. The latent
segment therefore adds internal computation without lengthening the visible
reasoning chain or repeatedly acquiring new image views. It provides an internal process in which question-relevant visual information
can continue to influence intermediate computation before the answer is committed
to language. We therefore refer to this process as latent visual reasoning in an operational rather than representational sense: the latent states are not assumed to constitute a visually disentangled space; instead, they provide additional multimodal computation whose dependence on question-relevant visual evidence is explicitly encouraged during training.

Additional latent computation alone, however, does not guarantee meaningful
visual reasoning. Without explicit visual supervision, the trajectory may
degenerate into arbitrary hidden computation that is useful only indirectly for
predicting the final answer. \textsc{MedLVR} therefore adopts a two-stage
learning strategy. Stage~1 biases the multimodal latent trajectory toward representations associated with question-relevant visual evidence. Stage~2 then optimizes whether the
resulting internal process contributes to a successful answer.

\subsection{Region-Supervised Latent Evidence Formation}
\label{sec:latent_evidence_formation}

The objective of Stage~1 is to give the latent trajectory an explicit
question-relevant visual target. The region annotation is used only as a
training signal and is never provided to the model during inference. Its role is not to prescribe an explicit visual action or produce a grounding output, but to provide a question-associated visual reference that regularizes the latent trajectory during training.

For each ROI-annotated training sample, we denote the image by $I$, the
question by $q$, and the associated bounding box by
$b=[x_1,y_1,x_2,y_2]$. After ViT patchification, the image is represented on
an $H_p\times W_p$ patch grid. We project $b$ onto this grid and collect all
covered patch indices in
$\mathcal I_{\mathrm{roi}}\subseteq\{1,\ldots,N\}$. The corresponding
projected visual tokens are
$\tilde V_{\mathrm{roi}}
=
\{\tilde v_j\mid j\in\mathcal I_{\mathrm{roi}}\}$.

Because the projected visual tokens $\tilde v_j$ and latent states $h_k$ lie
in the same $d$-dimensional model space, the ROI tokens can directly specify
the visual content expected to emerge from latent reasoning. We define the ROI
evidence target as
\begin{equation}
e^{\mathrm{roi}}
=
\frac{1}{|\mathcal I_{\mathrm{roi}}|}
\sum_{j\in\mathcal I_{\mathrm{roi}}}
\tilde v_j,
\qquad
e^{\mathrm{roi}}\in\mathbb{R}^{d},
\label{eq:roi_evidence_target}
\end{equation}
where $|\mathcal I_{\mathrm{roi}}|$ is the number of visual patches covered by
the annotated region.

The representation $e^{\mathrm{roi}}$ is not a localization prediction.
Neither the bounding box nor the pooled ROI representation is provided during
inference. Instead, $e^{\mathrm{roi}}$ serves as a latent evidence-formation
target that specifies the question-associated visual content toward which the
latent trajectory is encouraged to align.

This interpretation distinguishes \textsc{MedLVR} from explicit
region-revisiting approaches. The latter obtain evidence by searching for a
region, constructing a new view, and re-encoding that observation.
\textsc{MedLVR} keeps the original visual context unchanged and uses region
annotations only to shape internal evidence formation. The ROI therefore
specifies what evidence should emerge from reasoning rather than prescribing a
visual operation that must be executed.

\paragraph{Normalized representation geometry.}

The magnitude and coordinate scale of hidden representations can vary across
samples and training stages. Direct Euclidean distances may therefore reflect
changes in representation scale in addition to changes in content. For Stage~1
evidence formation and the representation-level analyses in
Section~\ref{sec:mechanistic_validation}, we use a parameter-free normalized
geometry. For any vector $z\in\mathbb{R}^{d}$, let
$\mu(z)=d^{-1}\sum_{r=1}^{d}z_r$ denote its feature mean and
$\nu(z)=d^{-1}\sum_{r=1}^{d}(z_r-\mu(z))^2$ its feature variance. We define
$\mathcal N(z)=(z-\mu(z))/\sqrt{\nu(z)+\epsilon_n}$, where
$\epsilon_n>0$ is a small numerical constant. The distance between two
representations is then
\begin{equation}
D(a,b)
=
\frac{1}{d}
\left\|
\mathcal N(a)-\mathcal N(b)
\right\|_2^2.
\label{eq:normalized_latent_distance}
\end{equation}
The factor $1/d$ normalizes the distance with respect to the hidden dimension.
Importantly, this normalized representation distance is used for Stage~1
evidence formation and the perturbation-based trajectory analysis in
Section~\ref{sec:mechanistic_validation}. The ROI-specific evidence-gap analysis
uses cosine similarity as a complementary measure of representational
selectivity. The Stage~2 latent policy is defined separately through an explicit
Gaussian likelihood in the original latent space.

\paragraph{Latent evidence formation.}

Given an image--question pair $(I,q)$, the model generates the trajectory
$H^{\mathrm{lat}}=\{h_k\}_{k=1}^{K}$. We constrain the trajectory using the
shared ROI evidence target,
\begin{equation}
\mathcal L_{\mathrm{form}}
=
\frac{1}{K}
\sum_{k=1}^{K}
D
\!\left(
h_k,
e^{\mathrm{roi}}
\right)
+
\gamma
D
\!\left(
h_K,
e^{\mathrm{roi}}
\right),
\label{eq:evidence_formation_loss}
\end{equation}
where $\gamma\geq0$ controls the additional supervision applied to the final
latent state.

The first term encourages the entire latent trajectory to remain aligned with
visual content associated with the current question. The second emphasizes $h_K$,
which has undergone the full latent reasoning budget and therefore represents
the most developed internal state before answer generation. This formulation
does not impose an arbitrary correspondence between latent-step index and
spatial patch order. Every latent state is treated as part of the same evidence
process, while the final state receives additional supervision because it
summarizes the complete latent segment.

The ROI therefore does not define a sequence of regions that the model must
visit. It defines the visual content that should remain available after internal
latent computation. Region supervision shapes latent evidence formation rather
than reproducing an external crop--reencode procedure inside the model.

\paragraph{Answer supervision.}

The latent trajectory must ultimately support correct answer generation. Given
the multimodal context $X_0$ and the latent trajectory
$H^{\mathrm{lat}}$, we optimize the standard next-token prediction objective
\begin{equation}
\mathcal L_{\mathrm{gen}}
=
-
\sum_{t=1}^{T_y}
\log
p_{\theta}
\!\left(
y_t
\mid
y_{<t},
X_0,
H^{\mathrm{lat}}
\right),
\label{eq:generation_loss}
\end{equation}
where $y_{<t}=\{y_1,\ldots,y_{t-1}\}$. The complete Stage~1 objective is
\begin{equation}
\mathcal L_{\mathrm{sft}}
=
\mathcal L_{\mathrm{gen}}
+
\lambda
\mathcal L_{\mathrm{form}},
\label{eq:sft_objective}
\end{equation}
where $\lambda\geq0$ balances answer generation and latent evidence formation.

Stage~1 therefore gives the continuous trajectory an explicit visual target.
The ROI specifies the visual content toward which the latent process is
encouraged to align, while the answer objective keeps the resulting
representation useful for medical VQA.

\subsection{Visual--Latent Policy Optimization}
\label{sec:vlpo}

Region-supervised training determines what visual evidence should be formed by
the latent trajectory, but it does not directly optimize whether a particular
internal trajectory leads to a better task outcome. Conventional outcome-level
optimization acts on generated text. Once intermediate reasoning is represented
by continuous latent states, the task reward should also influence the latent
decisions that precede answer generation.

We therefore introduce Visual--Latent Policy Optimization (VLPO), which extends
outcome-level learning from discrete textual actions to continuous latent
actions. The decoder transition is deterministic when its conditioning history
is fixed, whereas likelihood-ratio policy optimization requires an explicit
policy density. During Stage~2 training, VLPO treats each decoder-produced
latent state as the mean of a local stochastic policy. The sampled latent action
is then reused as the input embedding of the next latent step. This construction
gives each latent transition a well-defined likelihood while preserving the
autoregressive latent reasoning structure.

For each image--question pair $(I,q)$, we sample a group of $G$ trajectories
from a frozen behavior policy $\pi_{\theta_{\mathrm{old}}}$, obtaining
$\{y^{(i)},Z^{\mathrm{lat},(i)}\}_{i=1}^{G}$. Here, $i$ indexes a sampled
trajectory, $y^{(i)}$ is the generated textual sequence, and
$Z^{\mathrm{lat},(i)}=\{z_k^{(i)}\}_{k=1}^{K}$ is the corresponding sequence
of sampled latent actions.

Each trajectory receives an outcome reward
$R^{(i)}=R_{\mathrm{acc}}^{(i)}$, where
$R_{\mathrm{acc}}^{(i)}\in\{0,1\}$ indicates answer correctness. We compute
the group-normalized advantage
\begin{equation}
\hat A^{(i)}
=
\frac{
R^{(i)}
-
\mathrm{mean}\!\left(R^{(1)},\ldots,R^{(G)}\right)
}{
\mathrm{std}\!\left(R^{(1)},\ldots,R^{(G)}\right)
+
\epsilon_0
},
\label{eq:group_advantage}
\end{equation}
where $\epsilon_0>0$ is a small numerical constant. The same trajectory-level
advantage is assigned to all textual and latent actions belonging to trajectory
$i$, providing a shared sequence-level learning signal under sparse outcome
rewards.

\paragraph{Matched-context replay for textual actions.}

Likelihood ratios are meaningful only when the current and behavior policies
are evaluated under the same conditioning history. Otherwise, a change in token
probability could arise from a different latent trajectory rather than from a
change in the textual policy. We therefore replay each sampled trajectory and
insert the recorded latent actions $Z^{\mathrm{lat},(i)}$ at the corresponding
latent positions for both $\pi_{\theta}$ and
$\pi_{\theta_{\mathrm{old}}}$.

For sampled textual action $a_t^{\mathrm{txt},(i)}$, the importance ratio is
\begin{equation}
r_t^{\mathrm{txt}}(\theta)
=
\frac{
\pi_{\theta}
\!\left(
a_t^{\mathrm{txt},(i)}
\mid
s_t^{\mathrm{txt},(i)}
\right)
}{
\pi_{\theta_{\mathrm{old}}}
\!\left(
a_t^{\mathrm{txt},(i)}
\mid
s_t^{\mathrm{txt},(i)}
\right)
},
\label{eq:text_ratio}
\end{equation}
where $s_t^{\mathrm{txt},(i)}$ denotes the replayed multimodal, textual, and
latent history. Because both policies are evaluated under the same replayed
history, $r_t^{\mathrm{txt}}(\theta)$ isolates changes in textual likelihood
from changes in latent context.

\paragraph{Stochastic latent policy under matched-context replay.}

Let $\mu_k^{(i)}(\theta)\in\mathbb{R}^{d}$ denote the latent mean produced by
the current policy at step $k$ under matched-context replay. The replayed state
$s_k^{\mathrm{lat},(i)}$ contains the multimodal context $X_0$, the preceding
sampled text, and the recorded latent prefix
$\{z_1^{(i)},\ldots,z_{k-1}^{(i)}\}$.

During Stage~2 training, we define a local Gaussian policy
\begin{equation}
p_{\theta}
\!\left(
z_k^{(i)}
\mid
s_k^{\mathrm{lat},(i)}
\right)
=
\mathcal N
\!\left(
z_k^{(i)};
\mu_k^{(i)}(\theta),
\sigma^2 I
\right),
\label{eq:latent_policy}
\end{equation}
where $\sigma>0$ controls the exploration scale. The frozen behavior policy
samples
\begin{equation}
z_k^{(i)}
\sim
\mathcal N
\!\left(
\mu_k^{(i)}(\theta_{\mathrm{old}}),
\sigma^2 I
\right),
\label{eq:latent_sampling}
\end{equation}
and the sampled action $z_k^{(i)}$ is reused as the input embedding of the next
latent step.

The same sampled latent action is evaluated under the current and behavior
policies, giving
\begin{equation}
r_k^{\mathrm{lat}}(\theta)
=
\frac{
p_{\theta}
\!\left(
z_k^{(i)}
\mid
s_k^{\mathrm{lat},(i)}
\right)
}{
p_{\theta_{\mathrm{old}}}
\!\left(
z_k^{(i)}
\mid
s_k^{\mathrm{lat},(i)}
\right)
}.
\label{eq:latent_ratio}
\end{equation}
Because both policies use the same isotropic covariance, their Gaussian
normalization constants cancel and the ratio becomes
\begin{equation}
r_k^{\mathrm{lat}}(\theta)
=
\exp
\left[
-\frac{
\left\|
z_k^{(i)}
-
\mu_k^{(i)}(\theta)
\right\|_2^2
-
\left\|
z_k^{(i)}
-
\mu_k^{(i)}(\theta_{\mathrm{old}})
\right\|_2^2
}{
2\sigma^2
}
\right].
\label{eq:latent_ratio_expanded}
\end{equation}
Because the sampled action need not coincide with either policy mean,
$r_k^{\mathrm{lat}}(\theta)$ may be smaller or larger than one, reflecting
whether the current policy assigns lower or higher likelihood than the behavior
policy to the same latent action under an identical replayed history.

\paragraph{Factorized multimodal trajectory policy.}

A sampled \textsc{MedLVR} trajectory contains discrete textual actions and
continuous latent actions. Conditioned on the image--question pair, the
stochastic trajectory policy factorizes as
\begin{equation}
\pi_{\theta}(\tau\mid I,q)
=
\prod_{t=1}^{T_{\mathrm{txt}}}
\pi_{\theta}
\!\left(
a_t^{\mathrm{txt}}
\mid
s_t^{\mathrm{txt}}
\right)
\prod_{k=1}^{K}
p_{\theta}
\!\left(
z_k
\mid
s_k^{\mathrm{lat}}
\right),
\label{eq:joint_policy_factorization}
\end{equation}
where $T_{\mathrm{txt}}$ denotes the number of policy-sampled textual actions.

VLPO does not construct a single trajectory-level product of likelihood ratios,
which can become numerically unstable as multimodal trajectories grow longer.
Instead, it uses a factorized PPO-style surrogate: textual and latent actions
are evaluated under matched contexts, clipped separately, and aggregated with
length normalization. This enables position-wise likelihood-ratio optimization
under a shared trajectory-level advantage while avoiding the instability of
trajectory-level ratio products.

\paragraph{Joint clipped objective.}

For textual action $t$, we use the clipped surrogate
\begin{equation}
\ell_t^{\mathrm{txt}}
=
\min
\Big(
r_t^{\mathrm{txt}}(\theta)\hat A^{(i)},
\,
\mathrm{clip}
\!\left(
r_t^{\mathrm{txt}}(\theta),
1-\epsilon,
1+\epsilon
\right)
\hat A^{(i)}
\Big),
\label{eq:text_clipped_objective}
\end{equation}
and for latent action $k$,
\begin{equation}
\ell_k^{\mathrm{lat}}
=
\min
\Big(
r_k^{\mathrm{lat}}(\theta)\hat A^{(i)},
\,
\mathrm{clip}
\!\left(
r_k^{\mathrm{lat}}(\theta),
1-\epsilon,
1+\epsilon
\right)
\hat A^{(i)}
\Big),
\label{eq:latent_clipped_objective}
\end{equation}
where $\epsilon>0$ denotes the clipping range.

The complete VLPO objective is
\begin{equation}
\begin{aligned}
\mathcal J_{\mathrm{vlpo}}(\theta)
&=
\mathbb E_{(I,q)}
\,
\mathbb E_{\tau\sim\pi_{\theta_{\mathrm{old}}}}
\Bigg[
\frac{1}{T_{\mathrm{txt}}}
\sum_{t=1}^{T_{\mathrm{txt}}}
\ell_t^{\mathrm{txt}}
+
\lambda_{\mathrm{lat}}
\frac{1}{K}
\sum_{k=1}^{K}
\ell_k^{\mathrm{lat}}
\Bigg]
\\
&\quad
-
\beta
\,
\mathrm{KL}_{\mathrm{txt}}
\!\left(
\pi_{\theta}
\|
\pi_{\mathrm{ref}}
\right),
\end{aligned}
\label{eq:vlpo_objective}
\end{equation}
where $\lambda_{\mathrm{lat}}\geq0$ controls the contribution of latent-level
optimization, $\pi_{\mathrm{ref}}$ is a fixed reference policy, and
$\beta\geq0$ controls textual KL regularization.
$\mathrm{KL}_{\mathrm{txt}}$ denotes the expected token-distribution KL
divergence over replayed textual contexts in the optimization batch. The latent
policy is constrained through the clipped latent likelihood ratio. Length
normalization prevents the relative contribution of textual and latent actions
from depending on textual sequence length or latent budget.

Stage~1 determines what question-relevant visual information the latent
trajectory is encouraged to represent. VLPO then optimizes whether the sampled
internal trajectory contributes to a successful answer. Stochasticity is used only during Stage~2
training to define exploratory latent actions and their likelihood ratios. At
inference, we use the deterministic latent means $\mu_k(\theta)$ rather than
sampling, eliminating stochastic variation in the latent reasoning path while
preserving fixed-budget decoding.

\subsection{Inference and Optimization Details}
\label{sec:implementation_details}

\paragraph{Latent-segment decoding.}

At inference, \textsc{MedLVR} follows ordinary autoregressive decoding until
the latent segment is entered. The decoder then performs a fixed budget of $K$
latent updates. At each step, the final-layer hidden state is reused as the
input embedding of the next step. After the latent segment is completed,
ordinary token generation resumes.

The projected visual tokens $\tilde v_j$ and deterministic latent states $h_k$
both lie in $\mathbb{R}^{d}$, allowing region-derived visual evidence and
latent reasoning states to be compared within the same representation space.
We use a fixed latent budget rather than dynamic termination, keeping the
additional inference cost explicit and predictable.

\paragraph{Matched-context trajectory replay.}

For text positions, the recorded latent actions are inserted into the replayed
sequence for both $\pi_{\theta}$ and $\pi_{\theta_{\mathrm{old}}}$. Both
policies therefore evaluate each sampled textual action under identical visual,
textual, and latent context.

For latent step $k$, we replay the same multimodal context, preceding sampled
text, and recorded latent prefix
$\{z_1^{(i)},\ldots,z_{k-1}^{(i)}\}$. The current and behavior means,
$\mu_k^{(i)}(\theta)$ and
$\mu_k^{(i)}(\theta_{\mathrm{old}})$, are therefore evaluated under the same
state $s_k^{\mathrm{lat},(i)}$. The resulting likelihood ratio measures a
policy change at the current latent action rather than a discrepancy caused by
different preceding trajectories.

\paragraph{Optimization stability.}

Each sampled trajectory receives one sparse outcome reward, and the resulting
group-normalized advantage is shared by all textual and latent actions in that
trajectory. Textual and latent ratios are clipped separately using the same
range $\epsilon$. The coefficient $\beta$ controls token-level deviation from
the reference policy, while $\lambda_{\mathrm{lat}}$ controls the relative
weight of latent optimization and $\sigma$ sets the exploration scale of the
Gaussian latent policy.

In implementation, latent likelihood ratios are evaluated from differences of
Gaussian log likelihoods under matched replay contexts before conversion to the
ratio used by the clipped surrogate. The constants $\epsilon_0$ and
$\epsilon_n$ are used only for numerical stability in advantage normalization
and Stage~1 representation normalization, respectively.

The two training stages address complementary aspects of the proposed reasoning
process. Region-supervised learning shapes which question-associated visual
information is represented by the latent states. VLPO then optimizes whether the
resulting internal trajectory contributes to a successful answer.
\textsc{MedLVR} therefore first introduces a visual target for latent
computation and subsequently optimizes the task utility of the trajectory.

\FloatBarrier
\section{Experiments}
\label{sec:experiments}

\subsection{Experimental Setup}
\label{sec:experimental_setup}
\paragraph{Training data and protocol.}
\textsc{MedLVR} is trained in two stages using different data sources.
Stage~1 performs supervised fine-tuning on the large-scale medical grounding
corpus described in Section~\ref{sec:dataset}, which comprises approximately
1.2 million question--answer--box triplets across eight imaging modalities:
CT, chest X-ray, dermoscopy, endoscopy, fundus photography, MRI, OCT, and
ultrasound. This stage provides region-level supervision for learning the
visual evidence represented by the latent trajectory.

Stage~2 performs reinforcement learning on the open-access split of
OmniMedVQA, following the medical VQA protocol used in prior reasoning work
such as Med-R1. OmniMedVQA contains 82,059 medical images and 88,996
image--question pairs spanning eight evaluation modalities: CT, MRI, X-ray,
ultrasound, dermoscopy, fundus, OCT, and microscopy. We adopt an 80/20
train--test split, using the training portion for policy optimization and
reserving the test portion for in-domain evaluation.

\paragraph{External evaluation benchmarks.}
To evaluate transfer beyond the OmniMedVQA training distribution, we further
test \textsc{MedLVR} on five public medical VQA benchmarks: SLAKE
\cite{liu2021slake}, VQA-RAD \cite{lau2018dataset}, PMC-VQA
\cite{zhang2023pmc}, MMMU (Health \& Medicine) \cite{yue2024mmmu}, and
MedXpertQA \cite{zuo2025medxpertqa}. These benchmarks differ in image
characteristics and question formulation, while also covering substantially
different answer spaces. They therefore provide a complementary evaluation of
whether the learned latent reasoning process transfers beyond the data used for
policy optimization.

\paragraph{Evaluation protocol.}
We evaluate all model outputs using question-type-specific
answer matching. For binary questions, semantically equivalent
affirmative and negative expressions are mapped to canonical
positive and negative labels before exact matching. For
multiple-choice questions, predictions are mapped to the
corresponding answer option and evaluated using option-level
exact match. Open-ended responses are normalized by lowercasing,
removing punctuation, and collapsing whitespace before exact
matching. We do not use substring-based matching, as it may
produce clinically incorrect matches in the presence of negation,
laterality, or numerical disagreement, such as ``pneumothorax''
versus ``no pneumothorax.'' The same evaluation procedure is
applied to all models evaluated in our experiments.

\paragraph{Baseline models.}
We compare \textsc{MedLVR} with representative models spanning general-purpose
multimodal understanding and medical-domain adaptation. The general-purpose
group includes Gemini 2.5 Pro \cite{comanici2025gemini}, LLaVA-v1.5
\cite{liu2024improved}, and Qwen-series models \cite{yang2025qwen3}. The
medical-domain group includes Med-Flamingo \cite{moor2023med} and LLaVA-Med
\cite{li2023llava}.

We further compare against multimodal medical agent systems, including
MMedAgent \cite{li2024mmedagent} and VILA-M3 \cite{nath2025vila}, as well as
recent reasoning-oriented models such as MedVLM-R1 \cite{pan2025medvlm},
Pixel Reasoner \cite{wang2025pixel}, and MedVistaGym
\cite{lu2026medvistagym}. Together, these baselines cover text-based reasoning,
medical specialization, and methods that explicitly revisit visual evidence.

\paragraph{Baseline implementation and result sources.}
Baselines with publicly available model weights were re-evaluated
using our unified protocol, including the same test split, answer
normalization procedure, and decoding configuration.
For methods whose official code or model weights were unavailable
at the time of this study, we report the values provided in the
original publications, marked with $\dagger$ in
Tables~\ref{tab:modality_comparison} and~\ref{TABLE1}.
Because item-level predictions were unavailable for these methods,
they are included for contextual comparison only.
Our principal conclusions, including all ablation contrasts
(B1\textendash B7) and same-backbone improvements, are based
exclusively on models evaluated under an identical protocol on the
same test instances.

\begin{table*}[t]
\centering
\caption{Hyperparameter settings for Stage~1 and Stage~2 training.}
\label{tab:hyperparams}
\small
\renewcommand{\arraystretch}{1.12}
\setlength{\tabcolsep}{5pt}

\resizebox{\textwidth}{!}{%
\begin{tabular}{@{}llcc@{\qquad}llcc@{}}
\toprule
\textbf{Category} & \textbf{Parameter} & \textbf{Stage~1} & \textbf{Stage~2} &
\textbf{Category} & \textbf{Parameter} & \textbf{Stage~1} & \textbf{Stage~2} \\
\midrule

\multirow{5}{*}{Optimization}
& Learning rate      & $1\!\times\!10^{-5}$ & $2\!\times\!10^{-8}$ &
\multirow{5}{*}{VLPO}
& Group size $G$                         & -- & 4 \\
& Weight decay       & 0.1 & 0.1 &
& Exploration scale $\sigma$             & -- & 0.05 \\
& Warmup ratio       & 0.03 & 0.03 &
& Clipping range $\epsilon$              & -- & 0.2 \\
& LR schedule        & Cosine & Cosine &
& KL coefficient $\beta$                 & -- & 0.04 \\
& Training steps     & 2{,}500 & One epoch &
& Latent weight $\lambda_{\mathrm{lat}}$ & -- & 0.1 \\
\midrule

\multirow{4}{*}{\shortstack[l]{Latent\\reasoning}}
& Latent budget $K$             & 8 & 8 &
\multirow{4}{*}{Decoding}
& Temperature (training)        & -- & 0.2 \\
& Evidence weight $\lambda$     & 0.1 & -- &
& Maximum completion length     & 512 & 512 \\
& Final-state emphasis $\gamma$ & 1.0 & -- &
& Maximum prompt length         & 4096 & 4096 \\
& Evidence loss                 & MSE & -- &
& Inference decoding            & \multicolumn{2}{c}{Greedy} \\
\midrule

\multirow{2}{*}{Reward}
& Accuracy reward
& -- & \shortstack[c]{$\{0,1\}$, normalized\\exact matching} &
\multirow{2}{*}{Policy}
& Reference policy & -- & Fixed Stage~1 checkpoint \\
& Format reward
& -- & $\{0,0.2\}$ &
& Behavior policy & -- & Current policy (detached) \\
\midrule

\multicolumn{2}{@{}l}{Visual tokenization} &
\multicolumn{2}{c}{Minimum $128\!\times\!28\!\times\!28$} &
\multicolumn{2}{l}{Maximum visual tokenization} &
\multicolumn{2}{c}{Maximum $768\!\times\!28\!\times\!28$} \\

\bottomrule
\end{tabular}%
}
\end{table*}

\paragraph{Implementation details.}
All experiments are implemented in PyTorch and trained with DeepSpeed ZeRO-2
on a single server equipped with 6 NVIDIA A100 GPUs. We initialize
\textsc{MedLVR} from Qwen2.5-VL-7B-Instruct and perform full-parameter
optimization. During reinforcement learning, the vision tower and multimodal
merger remain frozen. Training uses bfloat16 mixed precision with gradient
checkpointing and FlashAttention-2. The per-device batch size is 4, and
gradient accumulation yields an effective global batch size of 48.
Stage~1 performs ROI-supervised fine-tuning to shape the visual evidence
represented by the latent trajectory. Stage~2 initializes from the resulting
model and applies Visual--Latent Policy Optimization (VLPO).
Table~\ref{tab:hyperparams} summarizes the complete hyperparameter settings
for both stages. Dynamic visual tokenization uses a minimum pixel budget of
$128\times28\times28$ and a maximum pixel budget of $768\times28\times28$.
Checkpoints are saved periodically, and data loading uses 8 workers.

The reference policy $\pi_{\mathrm{ref}}$ is initialized from the Stage~1
checkpoint and remains frozen throughout Stage~2. The behavior policy used
for importance ratio computation is the current policy with detached
gradients, updated at every optimization step.

The reward signal combines an accuracy reward
$R_{\mathrm{acc}} \in \{0, 1\}$ and a format reward
$R_{\mathrm{fmt}} \in \{0, 0.2\}$, summed with equal weight. The accuracy
reward is computed by exact string matching after answer extraction from
the first \texttt{<answer>...</answer>} block. Both predicted and reference
answers are lowercased with punctuation, quotation marks, and parentheses
removed and whitespace collapsed. A prediction is scored as correct if it
matches the reference exactly or, for answers of four or more characters,
if either string contains the other. The group-normalized advantage is
computed within each prompt's $G$ sampled trajectories as
$\hat{A}^{(i)} = (R^{(i)} - \bar{R}) / (\mathrm{std}(R) + 10^{-4})$.

The Gaussian log-likelihood under the latent policy
(Eq.~\ref{eq:latent_policy}) is computed as the sum over hidden dimensions.
To prevent numerical overflow in the importance ratio, we compute the ratio
in log space as the difference of log-likelihoods and clamp the resulting
log-ratio to $[-20, 20]$ before exponentiation. The exploration scale
$\sigma = 0.05$ is chosen to keep sampled latent actions close to the
deterministic decoder hidden states while providing sufficient exploration
for policy differentiation.

At inference, we use the deterministic latent means $\mu_k(\theta)$ rather
than sampling from the Gaussian policy, eliminating stochastic variation in
the latent reasoning path. Textual decoding uses greedy search. This
configuration is applied identically across all benchmarks.

\begin{table*}[tbp]
    \centering
    \caption{\textbf{Medical VQA performance across eight imaging modalities.}
    We compare \textsc{MedLVR} with zero-shot general-purpose VLMs,
    zero-shot medical VLMs, and recent reasoning-oriented medical VLMs.
    The evaluation covers CT, MRI, X-ray, Ultrasound, Dermoscopy, Fundus,
    OCT, and Microscopy. For \textsc{MedLVR}, all modality-wise scores are
    reported using a single fixed latent budget of $K=8$, without
    modality-specific tuning. Overall denotes accuracy weighted by the number
    of test samples in each modality.
    Results marked with $\dagger$ are quoted from the original publications;
    all unmarked results were obtained under our evaluation protocol.}
    \begin{adjustbox}{max width=\textwidth}
    {\footnotesize
    \begin{tabular}{l|cccccccc|c}
        \toprule
        \textbf{Methods} & \textbf{CT} & \textbf{MRI} & \textbf{X-Ray} &
        \textbf{Ultrasound} & \textbf{Dermoscopy} & \textbf{Fundus} &
        \textbf{OCT} & \textbf{Microscopy} & \textbf{Overall} \\
        \midrule
        \multicolumn{10}{c}{\textbf{Zero-shot VLMs}} \\
        \midrule
        BLIP-2$^\dagger$ & 56.7 & 41.3 & 67.6 & 37.3 & 40.7 & 46.2 & 68.1 & 50.4 & 48.18\\
        InstructBLIP$^\dagger$ & 28.7 & 33.2 & 61.0 & 41.3 & 62.2 & 50.3 & 42.6 & 46.3 & 40.35\\
        LLaVA$^\dagger$ & 17.7 & 26.7 & 30.7 & 18.7 & 49.7 & 47.1 & 33.7 & 28.9 & 27.92\\
        LLaMA Adapter v2$^\dagger$ & 21.4 & 26.6 & 46.4 & 34.1 & 51.8 & 50.7 & 33.0 & 38.7 & 32.77\\
        MiniGPT-4$^\dagger$ & 22.8 & 27.5 & 38.3 & 25.5 & 40.3 & 38.3 & 31.4 & 36.2 & 29.74\\
        InternVL2 & 40.2 & 58.1 & 57.9 & 49.1 & 51.9 & 53.2 & 59.1 & 64.0 & 53.40\\
        Qwen2-VL-2B & 45.1 & 38.6 & 39.3 & 30.9 & 35.8 & 43.2 & 35.1 & 36.9 & 38.76\\
        Qwen2-VL-7B & 61.5 & 45.8 & 64.3 & 36.0 & 49.1 & 59.8 & 59.3 & 61.1 & 51.95\\
        Qwen2-VL-72B & 68.0 & 69.4 & 77.2 & 51.4 & 65.3 & 72.6 & 72.8 & 67.8 & 67.70\\
        Qwen2.5-VL-3B & 53.9 & 54.2 & 61.8 & 32.7 & 52.9 & 62.5 & 56.2 & 59.6 & 53.17\\
        Qwen2.5-VL-7B & 60.4 & 58.4 & 74.0 & 30.7 & 62.5 & 67.3 & 61.2 & 67.8 & 58.52\\
        Qwen2.5-VL-72B & 66.2 & 68.7 & 77.6 & 49.8 & 69.8 & 71.0 & 69.2 & 69.4 & 67.13\\
        \midrule
        \multicolumn{10}{c}{\textbf{Zero-shot Medical VLMs}} \\
        \midrule
        LLaVA-Med$^\dagger$ & 18.7 & 27.5 & 30.7 & 29.9 & 45.0 & 39.0 & 34.6 & 33.3 & 29.17\\
        RadFM$^\dagger$ & 27.6 & 24.1 & 31.0 & 16.6 & 39.2 & 36.9 & 32.8 & 28.0 & 27.07\\
        Med-Flamingo$^\dagger$ & 38.5 & 40.6 & 30.3 & 24.6 & 32.4 & 30.1 & 26.5 & 19.9 & 34.16\\
        MedVInT$^\dagger$ & 40.7 & 43.1 & 55.1 & 41.3 & 29.1 & 31.8 & 23.3 & 32.0 & 40.16\\
        HuatuoGPT-Vision$^\dagger$ & 35.3 & 40.4 & 41.5 & 60.1 & 53.1 & 51.4 & 59.3 & 62.3 & 45.78\\
        HealthGPT$^\dagger$ & 35.5 & 78.5 & 81.9 & 51.4 & 64.9 & 54.6 & \textbf{89.3} & \textbf{88.2} & 66.37\\
        \midrule
        \multicolumn{10}{c}{\textbf{Thinking about Images}} \\
        \midrule
        Med-R1$^\dagger$ & 73.5 & 68.8 & \textbf{83.7} & 45.2 & 68.9 & 77.4 & 82.2 & 69.5 & 69.48\\
        \midrule
        \multicolumn{10}{c}{\textbf{Thinking with Images}} \\
        \midrule
        ViTAR$^\dagger$ & 64.0 & 68.6 & 81.0 & \textbf{84.9} & \textbf{82.3} & \textbf{83.4} & 84.2 & 78.4 & 74.10\\
        \midrule
        \multicolumn{10}{c}{\textbf{Thinking in Latent Visual Space}} \\
        \midrule
        \rowcolor{red!12}
        \textbf{\textsc{MedLVR} (ours)} & \textbf{80.4} & \textbf{82.0} & 77.5 & 51.8 & 76.6 & 81.9 & 80.2 & 70.2 & \textbf{76.52}\\
        \bottomrule
    \end{tabular}}
    \end{adjustbox}
    \label{tab:modality_comparison}
\end{table*}

\subsection{Medical VQA Performance across Imaging Modalities}
\label{sec:modality_results}

We first evaluate whether latent visual reasoning improves medical VQA across
heterogeneous imaging conditions. Table~\ref{tab:modality_comparison} reports
accuracy over eight modalities with substantially different visual
characteristics. Starting from the same Qwen2.5-VL-7B backbone,
\textsc{MedLVR} improves the overall accuracy from 58.52\% to 76.52\%. The gain
is observed across all eight modalities, with particularly large improvements
on CT (60.4\% to 80.4\%), MRI (58.4\% to 82.0\%), Ultrasound (30.7\% to
51.8\%), and OCT (61.2\% to 80.2\%). These results show that the benefit is
not confined to a single image domain or driven by one dominant modality.

Among the reasoning-oriented baselines, Med-R1 provides a relevant comparison
to the \emph{Thinking about Images} paradigm, in which intermediate reasoning is
performed in language space. Based on the values reported in the original
publication, \textsc{MedLVR} achieves a higher overall accuracy (76.52\% versus
69.48\%), with particularly large margins on CT and MRI of 6.9 and 13.2 points,
respectively. The model also performs strongly on Fundus and Dermoscopy,
reaching 81.9\% and 76.6\%. This pattern is consistent with the view that
additional latent computation can complement reasoning carried primarily
through textual trajectories, although as Med-R1 was not re-evaluated under
our protocol, this comparison should be interpreted as contextual rather
than strictly controlled.

ViTAR represents a different comparison point. As a \emph{Thinking with
Images} system, it relies on explicit visual interaction and is built on the
Lingshu backbone rather than the Qwen2.5-VL-7B backbone used by
\textsc{MedLVR}. The comparison is therefore not a controlled backbone-matched
ablation. Nevertheless, \textsc{MedLVR} achieves a comparable overall average
(76.52\% versus 74.10\%) without introducing external visual operations. The
relative advantage varies by modality: ViTAR remains substantially stronger on
Ultrasound and also leads on several other domains, whereas \textsc{MedLVR}
performs particularly well on CT and MRI. These differences indicate that the
two reasoning paradigms exhibit distinct modality-dependent strengths rather
than a uniform ordering across all visual domains.

Overall, the clearest evidence in Table~\ref{tab:modality_comparison} is the
consistent improvement over the same Qwen2.5-VL-7B backbone across all eight
modalities. The comparison with Med-R1 further shows an advantage over
text-space reasoning, while the contextual comparison with ViTAR suggests that internal
latent visual computation can remain competitive with a strong tool-based
visual reasoning system without requiring repeated external perception.

\begin{table*}[tbp]
    \centering
    \caption{\textbf{Performance comparison on external medical VQA benchmarks.}
    We compare \textsc{MedLVR} with proprietary frontier MLLMs, open-source
    general-purpose VLMs, medical MLLMs, multimodal medical agents, and
    recent reasoning-oriented multimodal models on SLAKE, VQA-RAD, PMC-VQA,
    MMMU (Health \& Medicine), and MedXpertQA.
    Bold and underlined values denote the best and second-best scores in each column.
    Superscripts in the \textsc{MedLVR} row report absolute improvements over Qwen2.5-VL-7B.
    Avg.\ is the arithmetic mean over all available benchmark scores in each row.
    Entries marked ``--'' are not included in the average.
    Results marked with $\dagger$ are quoted from the original publications
    because executable code or model weights were unavailable for re-evaluation;
    unmarked results were obtained by us using the same test split, answer
    normalization, and decoding configuration.
    Comparisons discussed in the text are based primarily on models
    we evaluated directly.}
    \label{TABLE1}

    \setlength{\tabcolsep}{5pt}
    \renewcommand{\arraystretch}{1.02}

    \begin{adjustbox}{max width=\textwidth}
    {\footnotesize
    \begin{tabular}{@{} l !{\sepvert} ccccc c @{}}
        \toprule
        \multirow{2}{*}{\textbf{Methods}} &
        \multicolumn{5}{c}{\textbf{Out-of-domain}} &
        \multirow{2}{*}{\textbf{Avg.}} \\
        \cmidrule(lr){2-6}
        & \textbf{SLAKE} & \textbf{VQA-RAD} & \textbf{PMC-VQA} &
          \textbf{MMMU(H\&M)} & \textbf{MedXpertQA} & \\
        \midrule

        \rowcolor{gray!12}
        \multicolumn{7}{@{}l}{\textbf{\textit{Closed-Source SOTA}}} \\
        \cmidrule(lr){1-7}
        GPT-4.1$^\dagger$        & 71.6 & 65.2 & 55.2 & \underline{73.6} & 40.8 & 61.3 \\
        GPT-5$^\dagger$          & \underline{73.2} & 64.5 & \underline{60.2} & 70.7 & 40.4 & 61.8 \\
        OpenAI-o3$^\dagger$      & \textbf{75.3} & 66.0 & \textbf{61.5} & \textbf{74.5} & \underline{44.1} & \textbf{64.3} \\
        Gemini 2.5 Pro$^\dagger$ & 72.7 & 63.8 & 55.9 & 72.8 & \textbf{46.6} & \underline{62.4} \\
        \midrule

        \rowcolor{gray!12}
        \multicolumn{7}{@{}l}{\textbf{\textit{Open-Source SOTA}}} \\
        \cmidrule(lr){1-7}
        LLaVA-v1.5-8B$^\dagger$  & 59.4 & 54.2 & 36.4 & 38.2 & --   & -- \\
        InternVL3-8B$^\dagger$   & 70.4 & 65.6 & 52.7 & 62.3 & 23.8 & 55.0 \\
        LLaVA-Next-7B$^\dagger$  & 57.9 & 52.6 & 35.5 & 33.1 & 20.7 & 40.0 \\
        LLaVA-Next-13B$^\dagger$ & 57.1 & 54.8 & 36.6 & 40.1 & 19.6 & 41.6 \\
        Qwen2.5-VL-32B           & 70.1 & \textbf{71.7} & 50.4 & 60.1 & 26.8 & 55.8 \\
        \midrule

        \rowcolor{gray!12}
        \multicolumn{7}{@{}l}{\textbf{\textit{Medical MLLMs}}} \\
        \cmidrule(lr){1-7}
        Med-Flamingo$^\dagger$         & 43.5 & 45.4 & 23.3 & 28.3 & 19.3 & 32.0 \\
        RadFM$^\dagger$                & 34.6 & 50.6 & 25.9 & 27.0 & 19.8 & 31.6 \\
        LLaVA-Med-7B$^\dagger$         & 47.7 & 52.5 & 24.7 & 38.8 & 19.9 & 36.7 \\
        HuatuoGPT-Vision-34B$^\dagger$ & 68.3 & 61.7 & 51.4 & 60.1 & 23.6 & 53.0 \\
        \midrule

        \rowcolor{gray!12}
        \multicolumn{7}{@{}l}{\textbf{\textit{Multimodal Medical Agents}}} \\
        \cmidrule(lr){1-7}
        MMedAgent-7B$^\dagger$    & 68.7 & 64.0 & -- & 44.1 & 22.3 & -- \\
        AURA$^\dagger$            & 68.4 & 64.5 & -- & 49.3 & 23.5 & -- \\
        SMR-Agents$^\dagger$      & 53.5 & 46.9 & -- & 40.1 & 19.6 & -- \\
        MedAgent-Pro$^\dagger$    & 69.4 & 63.3 & -- & 52.9 & 27.8 & -- \\
        MMedAgent-RL-7B$^\dagger$ & 67.9 & \underline{66.1} & -- & 58.9 & 22.6 & -- \\
        VILA-M3-40B$^\dagger$     & 71.4 & 65.7 & -- & 56.6 & 23.0 & -- \\
        \midrule

        \rowcolor{gray!12}
        \multicolumn{7}{@{}l}{\textbf{\textit{Thinking about Images}}} \\
        \cmidrule(lr){1-7}
        MedVLM-R1$^\dagger$ & 58.9 & 45.2 & 44.8 & 45.9 & 21.7 & 43.3 \\
        Med-R1$^\dagger$    & 55.1 & 36.5 & 45.8 & 44.7 & 22.9 & 41.0 \\
        \midrule

        \rowcolor{gray!12}
        \multicolumn{7}{@{}l}{\textbf{\textit{Thinking with Images}}} \\
        \cmidrule(lr){1-7}
        DeepEyes-7B$^\dagger$              & 68.2 & 65.9 & --   & 57.8 & 23.6 & -- \\
        Mini-o3-7B-v1$^\dagger$            & 67.8 & 65.7 & --   & 57.4 & \textbf{24.3} & -- \\
        PixelReasoner-RL-v1-7B$^\dagger$   & 67.3 & 66.0 & --   & 58.0 & 23.5 & -- \\
        MEDVISTAGYM (Qwen3vl-8B)$^\dagger$ & 27.0 & 51.9 & 49.5 & 42.9 & --   & -- \\
        \midrule

        \rowcolor{gray!12}
        Qwen2.5-VL-7B & 63.7 & 59.9 & 49.0 & 46.4 & 22.5 & 48.3 \\
        \rowcolor{red!12}
        MedLVR (ours) &
            66.4\textsuperscript{\textcolor{red!65!black}{(+2.7)}} &
            65.9\textsuperscript{\textcolor{red!65!black}{(+6.0)}} &
            53.6\textsuperscript{\textcolor{red!65!black}{(+4.6)}} &
            56.6\textsuperscript{\textcolor{red!65!black}{(+10.2)}} &
            24.3\textsuperscript{\textcolor{red!65!black}{(+1.8)}} &
            53.4 \\
        \bottomrule
    \end{tabular}}
    \end{adjustbox}
\end{table*}
\subsection{Generalization to External Medical VQA Benchmarks}
\label{sec:external_generalization}

We next evaluate whether the learned latent reasoning pathway transfers beyond
the distribution used for policy optimization. Table~\ref{TABLE1} reports
results on five external medical VQA benchmarks that differ in image sources,
question formats, and answer spaces. Starting from the same
Qwen2.5-VL-7B backbone, \textsc{MedLVR} improves performance on every
benchmark, increasing the average score from 48.3\% to 53.4\%. The largest
gain is observed on MMMU (Health \& Medicine), where accuracy increases from
46.4\% to 56.6\%. Consistent improvements are also obtained on VQA-RAD
(59.9\% to 65.9\%) and PMC-VQA (49.0\% to 53.6\%), while smaller gains are
observed on SLAKE and MedXpertQA.

The magnitude of the improvement varies across benchmarks rather than appearing
as a uniform offset. Nevertheless, the consistent gain over the same backbone
across all five datasets shows that the learned reasoning pathway remains useful
under changes in image distribution and question formulation. This result is
particularly important because Stage~2 policy optimization is performed only on
OmniMedVQA, whereas no additional adaptation is conducted on any of the five
external benchmarks.

The broader comparison in Table~\ref{TABLE1} provides additional context.
Larger open-source and proprietary models remain stronger on several datasets,
but \textsc{MedLVR} is competitive with recent reasoning-oriented medical
systems while using a single end-to-end inference procedure. Unlike methods
that rely on retrieval augmentation or repeated external visual operations,
\textsc{MedLVR} obtains its improvement by modifying the model's internal
reasoning process. The consistent same-backbone gains therefore provide the
clearest evidence that the proposed latent visual reasoning pathway improves
transfer beyond the distribution used for policy optimization.
Comparisons with other methods in Table~\ref{TABLE1} that were not
re-evaluated under our protocol are included for contextual positioning
and should be interpreted accordingly.


\subsection{Mechanistic Validation of Latent Visual Reasoning}
\label{sec:mechanistic_validation}

The preceding results show that \textsc{MedLVR} improves medical VQA
performance across imaging modalities and external benchmarks. Final task
accuracy, however, does not reveal what information is represented within the
latent trajectory or whether the resulting internal states are functionally
linked to the prediction. We therefore examine the proposed Stage~1 mechanism
through matched representational and intervention analyses, and then test
whether the full model remains dependent on visual input.

For ROI-based analysis, we construct a separate held-out mechanism set of 200
cases. This subset is intended for controlled intervention rather than
benchmark-level performance estimation. We retain only cases for which a
question-specific expert ROI is available and can be selectively removed
without altering the question itself. All cases are drawn from public sources
used to build the Stage~1 grounding corpus, but are excluded from both Stage~1
training and Stage~2 policy optimization. For each expert ROI, we sample
$R=5$ non-overlapping random regions with matched area, providing a control for
the generic effect of removing image content.

\begin{figure*}[tbp]
    \centering
    \includegraphics[width=\textwidth]{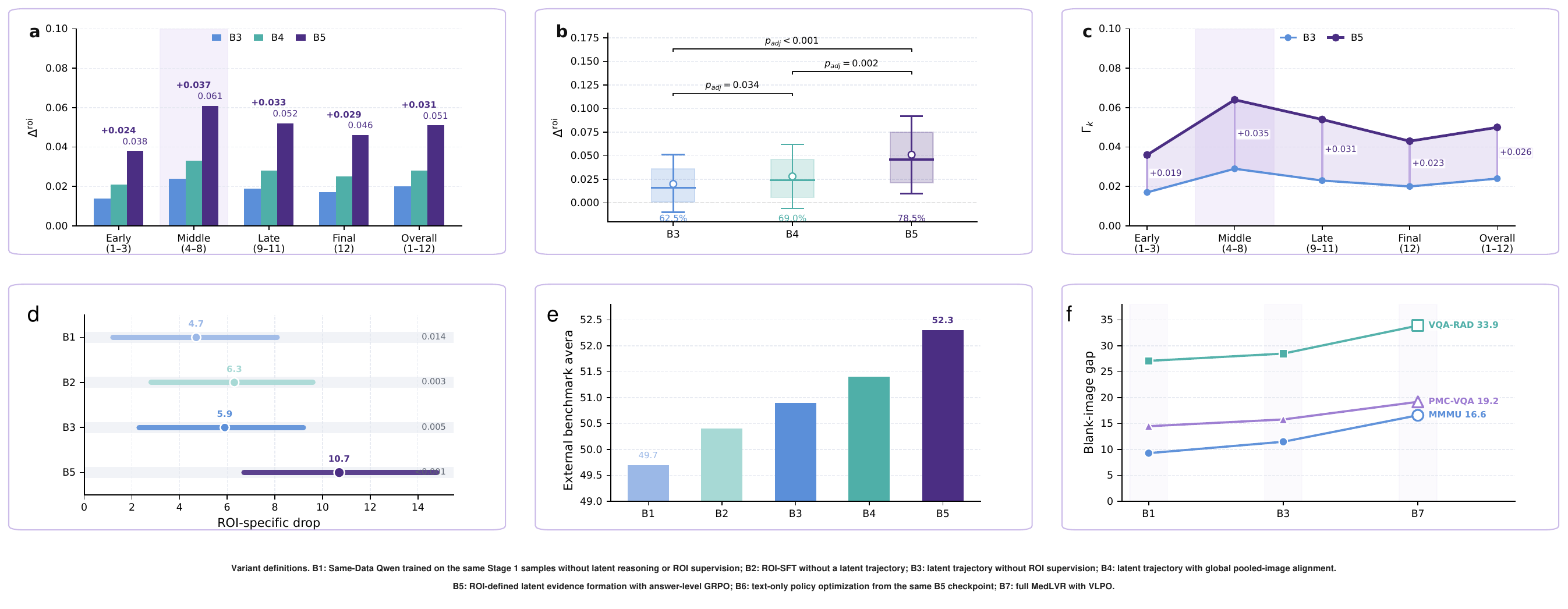}
    \caption{\textbf{Mechanistic validation of latent visual reasoning.}
    \textbf{a}, Stage-wise ROI-specific evidence gap for B3, B4, and B5.
    Latent steps are grouped into Early (1--3), Middle (4--8), Late (9--11),
    and Final (12) stages; Overall averages all 12 steps.
    \textbf{b}, Sample-level distribution of the overall ROI-specific evidence
    gap on the 200-case held-out mechanism set. Boxes denote the interquartile
    range, horizontal lines denote the median, circles denote the mean, and
    whiskers denote mean $\pm$ s.d. Percentages below the axis indicate the
    fraction of samples with a positive gap. Adjusted $p$ values are reported
    for pairwise comparisons.
    \textbf{c}, Stage-wise ROI perturbation sensitivity for B3 and B5.
    \textbf{d}, ROI-specific accuracy drop after masking the clinically
    relevant region. Points denote the estimated drop and horizontal intervals
    denote 95\% confidence intervals.
    \textbf{e}, Average accuracy across the five external medical VQA
    benchmarks for matched B1--B5 controls.
    \textbf{f}, Blank-image gap on MMMU (Health \& Medicine), VQA-RAD, and
    PMC-VQA for B1, B3, and B7.}
    \label{fig:mechanism_validation}
\end{figure*}

For trajectory-resolved mechanism analysis, we use a diagnostic latent budget
of $K=12$ to obtain finer temporal resolution across the latent reasoning
process. The trained checkpoints are kept unchanged, and the latent rollout is
extended from the default $K=8$ to $K=12$ only during this diagnostic
evaluation. All trajectory-based comparisons in
Fig.~\ref{fig:mechanism_validation} use the same $K=12$ configuration, whereas
the main benchmark results reported elsewhere use the default latent budget of
$K=8$.

Throughout the mechanism analysis, we use seven matched variants. B1 is the
\emph{Same-Data Qwen} baseline, trained on the same Stage~1
image--question--answer samples without latent reasoning or ROI supervision.
B2 adds ROI-SFT without a latent trajectory. B3 introduces latent reasoning
without ROI supervision. B4 replaces the ROI target with global pooled-image
alignment. B5 uses ROI-defined latent evidence formation with answer-level
GRPO. B6 starts from the same B5 checkpoint and applies text-only policy
optimization. B7 is the full \textsc{MedLVR} model with VLPO. These variants
separate additional medical data, ROI supervision, generic latent computation,
and direct optimization of the latent trajectory.

Figure~\ref{fig:mechanism_validation} summarizes the full analysis. Panels a--c characterize the visual selectivity and perturbation sensitivity
of the Stage~1 latent trajectory. Panels d and e test prediction-level dependence
and matched-data alternatives, while panel f examines whether the full model's
external-benchmark gains remain dependent on visual input.

\subsubsection{Formation of question-associated visual evidence in the latent trajectory}

We first test whether latent states preferentially represent the expert-defined
medical evidence required by the current question. For sample $i$ at latent
step $k$, we compare the hidden state $h_{i,k}$ with the representation of the
expert ROI, $e_i^{\mathrm{roi}}$, and with the representations of the matched
random regions, $e_{i,r}^{\mathrm{rand}}$. The ROI-specific evidence gap is
defined as $\Delta_{i,k}^{\mathrm{roi}}=
\operatorname{CosSim}(h_{i,k},e_i^{\mathrm{roi}})
-\frac{1}{R}\sum_{r=1}^{R}
\operatorname{CosSim}(h_{i,k},e_{i,r}^{\mathrm{rand}})$, where $R=5$.
A positive value indicates that the latent state is more closely aligned with
the expert-defined, question-associated ROI than with equally sized control
regions.

As shown in Fig.~\ref{fig:mechanism_validation}(a), B3, the latent-only control
without ROI supervision, exhibits only a modest ROI preference, with an overall
evidence gap of 0.020. B4, the global-alignment control, increases the value to 0.028, indicating that generic image-level
supervision increases visual selectivity in hidden-space computation. The effect is substantially stronger for B5, which uses ROI-defined latent
evidence formation and reaches an overall gap of 0.051.

The separation is observed throughout the trajectory and is largest in the
middle stage. B5 reaches 0.038, 0.061, 0.052, and 0.046 in the Early, Middle,
Late, and Final stages, respectively. Relative to B3, the corresponding gains
are 0.024, 0.037, 0.033, and 0.029. The peak therefore emerges after several
latent updates rather than at trajectory initialization. Evidence specificity
then remains elevated through the final state, suggesting that question-relevant
visual information remains represented after the middle-stage increase.

The sample-level analysis in Fig.~\ref{fig:mechanism_validation}(b) shows that
this effect is not driven by a small number of cases. The mean ROI-specific gap
increases from $0.020\pm0.031$ for B3 to $0.028\pm0.034$ for B4 and
$0.051\pm0.041$ for B5. The corresponding medians are 0.016, 0.024, and
0.046. Positive ROI specificity is observed in 62.5\% of B3 samples, 69.0\%
of B4 samples, and 78.5\% of B5 samples. Adjusted pairwise comparisons remain
significant for B3 versus B4 ($p_{\mathrm{adj}}=0.034$), B4 versus B5
($p_{\mathrm{adj}}=0.002$), and B3 versus B5
($p_{\mathrm{adj}}<0.001$). ROI-defined supervision therefore shifts the
distribution of latent evidence across the held-out set rather than increasing
the mean through a small number of extreme examples.

Panels a and b also distinguish the proposed mechanism from two simpler
alternatives. Additional latent steps alone produce only weak ROI specificity,
while global image alignment provides an intermediate effect. The strongest
evidence selectivity appears only when the latent trajectory is explicitly
shaped around the medically relevant region defined by the question.

\subsubsection{Does removing the true visual evidence alter the latent trajectory?}

Representational alignment alone does not establish that ROI information is
functionally involved in the internal computation. We therefore intervene on
the image and measure how the latent trajectory changes when the true ROI,
rather than a matched random region, is removed. For latent step $k$, we define
$\Gamma_k=
D(h_k(I),h_k(I^{-\mathrm{ROI}}))
-D(h_k(I),h_k(I^{-\mathrm{Rand}}))$, where
$I^{-\mathrm{ROI}}$ denotes the image with the expert ROI masked,
$I^{-\mathrm{Rand}}$ masks an area-matched random region, and $D$ is the same
representation-distance function for all variants. A larger $\Gamma_k$
indicates that removing the expert-defined region changes the latent state more
than generic image corruption of the same size.

Figure~\ref{fig:mechanism_validation}(c) shows that B3, the latent-only control,
remains only weakly sensitive to ROI removal, with an overall value of 0.024.
B5, the ROI-formation variant, reaches 0.050.
The largest separation again occurs in the middle stage, where sensitivity
increases from 0.029 for B3 to 0.064 for B5. B5 remains more sensitive in the
Late stage (0.054 versus 0.023) and at the Final state (0.043 versus 0.020).

The correspondence between panels a and c is central to the mechanism
analysis. The states that become more selective for the expert ROI are also the
states that change more when that ROI is removed. This correspondence is consistent with functional involvement of ROI-associated
information in the latent computation: removing the expert region produces a
larger trajectory change than removing an area-matched control region.

\subsubsection{Does question-associated visual evidence affect the final prediction?}

We next test whether the trajectory-level effect propagates to the answer. Each
held-out ROI case is evaluated under three otherwise identical conditions: the
original image, the image with the expert ROI masked, and the image with an
area-matched random region masked. The question, latent budget, and decoding
settings are fixed across conditions. We define
$\Delta_{\mathrm{ROI}}=\operatorname{Acc}(I)
-\operatorname{Acc}(I^{-\mathrm{ROI}})$ and
$\Delta_{\mathrm{Rand}}=\operatorname{Acc}(I)
-\operatorname{Acc}(I^{-\mathrm{Rand}})$, and measure selective dependence on
the relevant region as
$\Delta_{\mathrm{specific}}=\Delta_{\mathrm{ROI}}-\Delta_{\mathrm{Rand}}$.

Figure~\ref{fig:mechanism_validation}(d) shows that all matched variants depend
more strongly on the expert ROI than on a random region, but the magnitude of
this dependence differs substantially. B1, the same-data baseline, exhibits an ROI-specific drop of
4.7 points (95\% CI, 1.2--8.1; $p=0.014$). B2, which adds ROI-SFT without a latent trajectory, reaches 6.3 points
(95\% CI, 2.8--9.6; $p=0.003$), while B3, which adds latent reasoning without ROI supervision, reaches 5.9 points
(95\% CI, 2.3--9.2; $p=0.005$). B5, which combines the latent pathway with ROI-defined evidence formation, shows the largest effect, with a
10.7-point drop (95\% CI, 6.7--14.8; $p<0.001$).

Because the intervention includes an area-matched random-mask control, the
reported quantity reduces the influence of generic image corruption and
emphasizes the additional effect of removing the expert-defined region. B5 is therefore not simply more sensitive to image
corruption; its prediction depends more selectively on the visual evidence
identified by the expert ROI. The mechanism observed inside the trajectory
thus extends to the final answer: stronger ROI representation is accompanied
by a larger trajectory change under ROI removal and a larger prediction-level
effect.

\subsubsection{What explains the performance gain?}

A remaining alternative is that the final improvement arises from additional
medical data, ordinary ROI supervision, or generic hidden-space computation.
Figure~\ref{fig:mechanism_validation}(e) addresses these explanations with
matched-data controls. All B1--B5 variants use the same Stage~1 samples, the
same backbone, and the same training budget. The models differ only in whether
they include ROI supervision, a latent trajectory, or the form of visual target
used to shape that trajectory.

B1, the same-data baseline, reaches an external-benchmark average of 49.7. B2,
which adds ROI-SFT without a latent trajectory, increases the score to 50.4. B3,
which introduces latent reasoning without ROI supervision, raises the average to
50.9. B4, the global-alignment control, provides a further increase to 51.4.
B5, which combines the latent pathway with ROI-defined evidence formation,
reaches the highest average of 52.3.

The matched contrasts reveal an interaction that is not visible from the final
score alone. Without a latent trajectory, ROI supervision improves the average
by 0.7 points (B2 versus B1). Within the latent pathway, replacing no ROI
supervision with ROI-defined evidence formation improves the average by
1.4 points (B5 versus B3). Region supervision is therefore more effective when it shapes an internal
latent trajectory than when it is applied without one. Conversely, latent computation alone is useful, but generic global
alignment does not match a question-specific ROI target. These matched contrasts support a contribution from the coupling between latent
computation and region-defined evidence formation, rather than from any single
ingredient in isolation.

\subsubsection{Does the final gain depend on visual input?}

We finally test whether the external-benchmark improvement remains dependent on
the image itself. We evaluate B1 (Same-Data Qwen), B3 (latent reasoning without ROI supervision),
and B7 (full \textsc{MedLVR} with VLPO) on MMMU (Health \& Medicine),
VQA-RAD, and PMC-VQA using the original image and a blank-image intervention.
For each model, we define the blank-image gap as
$G_{\mathrm{blank}}=\operatorname{Acc}(\mathrm{Full})
-\operatorname{Acc}(\mathrm{Blank})$. A larger gap indicates that performance
degrades more strongly when visual content is removed.

As shown in Fig.~\ref{fig:mechanism_validation}(f), the blank-image gap
increases consistently from B1 to B3 and then to B7. On MMMU, the gap rises
from 9.3 to 11.5 and 16.6 points. On VQA-RAD, it increases from 27.1 to 28.5
and 33.9 points. PMC-VQA shows the same ordering, with gaps of 14.5, 15.8,
and 19.2 points.

The same ordering across three external benchmarks is inconsistent with a
purely text-only explanation for the gain. The full model benefits more from the
presence of visual input and loses more when that input is removed. Taken
together, panels a--f show that ROI-supervised latent states are more selective
for the expert region, are more sensitive to removing that region, and yield
predictions with stronger dependence on visual input. Matched-data controls
further show that additional samples, ordinary ROI-SFT, and generic latent
computation do not fully account for the observed effect. These findings are
consistent with the proposed mechanism of strengthening question-associated
visual information within the latent process.

\subsection{Direct Optimization of the Latent Reasoning Process}
\label{sec:latent_optimization}

The mechanism analysis above shows that ROI-supervised Stage~1 changes the
visual evidence represented and used by the latent trajectory. It does not
isolate the value of directly optimizing that trajectory during Stage~2. A
possible alternative is that the gain of VLPO comes from a stronger
answer-level policy optimizer rather than from optimization of the latent
reasoning process itself.

We therefore start from the same B5 checkpoint and construct three Stage~2
variants. B5, the ROI-formation model, uses standard answer-level GRPO. B6,
the text-only policy-optimization control, uses the same rollout
sampling, reward function, clipping, KL regularization, and optimization budget
as VLPO, but updates only the textual policy. B7 adds the latent-trajectory objective and corresponds to the full
\textsc{MedLVR} model with VLPO. Because B6 and B7 share the
same Stage~1 initialization and the same answer-level optimization framework,
their difference isolates the contribution of direct latent-level optimization.

Text-only policy optimization already improves the external-benchmark average
from 52.3 for B5 to 52.7 for B6, showing that the revised policy-optimization
framework contributes independently of the latent objective. Full VLPO further
raises the average to 53.4 and outperforms B6 on all five external benchmarks.
The remaining gain is therefore not attributable to additional data or a
stronger answer-level reward signal.

The largest separation occurs on MMMU (Health \& Medicine), where B6 reaches
55.3 and B7 reaches 56.6, a further improvement of 1.3 points. The gain is
smaller on SLAKE and MedXpertQA. This variation is consistent with the role of
the latent objective: directly shaping the internal trajectory is most useful
when the task requires sustained interaction with visual evidence before the
answer is formed.

The comparison between B6 and B7 provides the cleanest isolation of VLPO.
Both models start from the same ROI-formed trajectory and use the same Stage~2
optimization machinery. The only additional signal in B7 acts on the latent
states themselves. Its consistent advantage therefore indicates that
answer-level policy learning does not fully determine the quality of the
internal reasoning process. Direct optimization of the latent trajectory
provides an additional benefit beyond optimization of the final textual action.

Section~\ref{sec:mechanistic_validation} and the B6--B7 comparison define
distinct roles for the two training stages. Stage~1 determines what visual
evidence the latent trajectory is encouraged to form. Stage~2 determines
whether the resulting internal process is directly optimized for the task
outcome. ROI-defined evidence formation gives the latent computation a
question-relevant visual target, while VLPO makes the trajectory itself
responsive to downstream reward.

\par\medskip
\begin{center}
    \includegraphics[width=0.96\columnwidth]{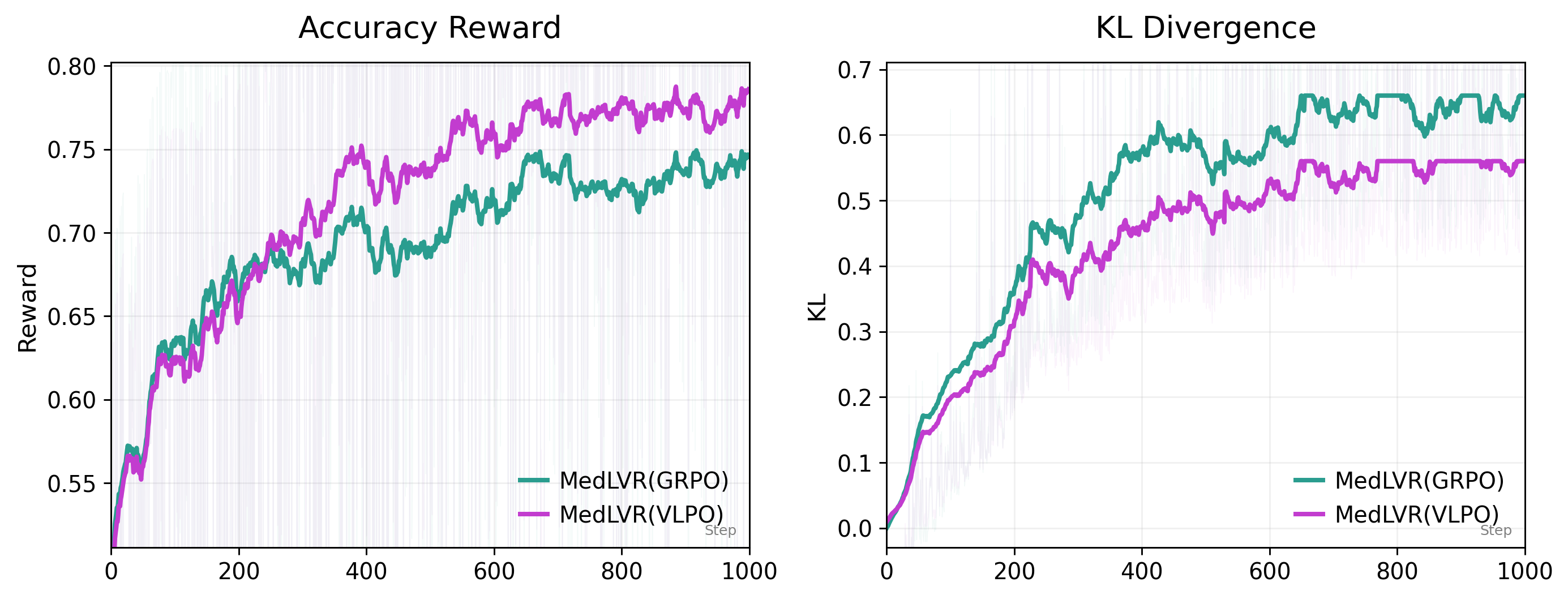}
    \captionof{figure}{\textbf{Optimization dynamics of GRPO and VLPO.}
    Left, accuracy reward over training steps. Right, KL divergence to the
    reference policy. VLPO maintains a higher accuracy reward through most of
    training while exhibiting lower KL divergence.}
    \label{fig:vlpo_grpo_dynamics}
\end{center}
\medskip

\subsection{Benchmark-Level Contribution of the Main Components}
\label{sec:component_analysis}

Sections~\ref{sec:mechanistic_validation} and the preceding VLPO analysis
isolate the functional roles of latent reasoning, ROI-defined evidence
formation, and direct latent-level optimization. We next examine whether the
resulting gains are consistent across external benchmarks. As shown in
Table~\ref{tab:variant_datasets}, performance improves progressively from B1
to B3, from B3 to B5, and finally from B5 to B7 across all five datasets.

\begin{table}[tbp]
\centering
\caption{\textbf{Contribution of the main training components.}
Performance of different reasoning and optimization variants across five
medical VQA benchmarks.}
\label{tab:variant_datasets}
\resizebox{\columnwidth}{!}{
\begin{tabular}{lccccc}
\toprule
\textbf{Variant} & \textbf{SLAKE} & \textbf{VQA-RAD} &
\textbf{PMC-VQA} & \textbf{MMMU-HM} & \textbf{MedXpertQA} \\
\midrule
B1: Same-Data Qwen                & 63.2 & 60.7 & 49.6 & 48.1 & 22.7 \\
B3: Latent SFT, no ROI            & 64.9 & 62.5 & 51.3 & 52.5 & 23.2 \\
B5: MedLVR + ROI Formation        & 65.7 & 64.7 & 52.8 & 54.7 & 23.8 \\
B6: Text-only Policy Optimization & 66.0 & 65.1 & 53.1 & 55.3 & 24.0 \\
B7: Full MedLVR                   & 66.4 & 65.9 & 53.6 & 56.6 & 24.3 \\
\bottomrule
\end{tabular}
}
\end{table}

Introducing the latent reasoning pathway without ROI supervision improves
performance over the same-data baseline on every benchmark. The largest gain
occurs on MMMU-HM, where accuracy increases from 48.1\% for B1 to 52.5\% for
B3. Adding ROI-defined evidence formation produces a further improvement on
all five datasets, raising MMMU-HM to 54.7\% and VQA-RAD from 62.5\% to
64.7\%. The full model with VLPO achieves the strongest result in every
column, including 56.6\% on MMMU-HM and 65.9\% on VQA-RAD.

The consistent ordering across datasets indicates that the contribution of the
main components is not driven by a single benchmark. Latent reasoning provides
the first improvement, ROI-defined evidence formation strengthens the learned
trajectory, and VLPO provides the final gain. Table~\ref{tab:variant_datasets}
therefore complements the controlled analyses above by showing that this
progression is preserved across heterogeneous external evaluation settings.

\subsection{Analysis of Visual Evidence Allocation}
\label{sec:visual_evidence_analysis}

\begin{figure*}[!t]
    \centering
    \includegraphics[width=\textwidth]{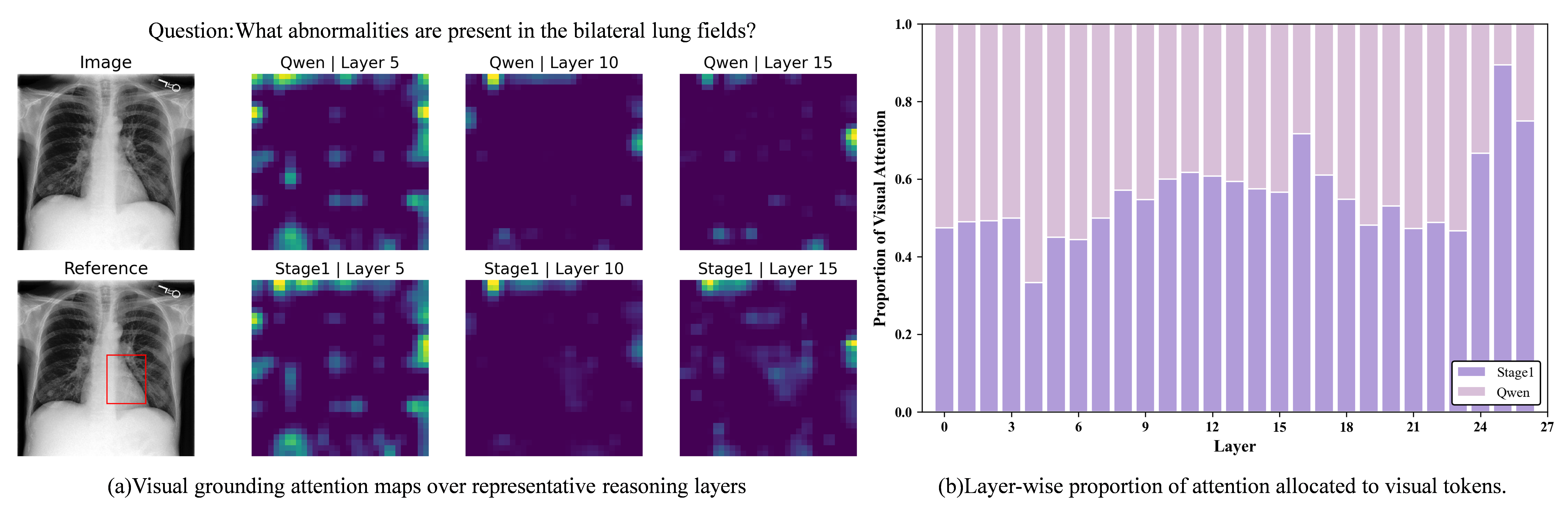}
    \caption{\textbf{Visual evidence allocation during reasoning.}
    Spatial activation maps compare the base model and \textsc{MedLVR} across
    decoder layers, while the layer-wise statistics quantify the relative
    allocation of attention to visual tokens.}
    \label{fig:attention_analysis}
\end{figure*}
Fig.~\ref{fig:attention_analysis} provides a complementary view of how
visual information is allocated during inference at spatial and layer-wise
levels. As shown in Fig.~\ref{fig:attention_analysis}(a), the base Qwen model
exhibits relatively diffuse activation over the image, whereas
\textsc{MedLVR} shows a more concentrated response around the clinically
relevant lung region. The difference becomes more apparent in intermediate
and deeper decoder layers, particularly at Layers 10 and 15.

The layer-wise statistics in Fig.~\ref{fig:attention_analysis}(b) show a
consistent pattern. \textsc{MedLVR} assigns a larger proportion of attention
to visual tokens across most decoder layers, with the separation becoming more
pronounced at greater depth. These observations suggest that the proposed model allocates a larger share of
attention to visual representations as decoding progresses.

This analysis is descriptive rather than causal. The attention maps do not by
themselves establish that the highlighted visual information determines the
final prediction. Instead, they complement the controlled interventions in
Section~\ref{sec:mechanistic_validation}, which directly test whether
question-relevant visual evidence alters the latent trajectory and affects the
answer.

\begin{table}[tbp]
\centering
\caption{\textbf{Inference efficiency and accuracy.}
Avg.\ Tokens denotes the average number of generated output tokens per sample,
and Total Time denotes total inference time on the 485-sample evaluation set.}
\label{tab:efficiency_accuracy}
\resizebox{\columnwidth}{!}{
\begin{tabular}{lccc}
\toprule
Method & Avg. Tokens & Total Time & Overall Accuracy \\
\midrule
MedGemma-1.5     & 102.1 & 1300.2s & 9.1\% \\
Chiron-o1        & 16.6  & 147.4s  & 44.3\% \\
Qwen2.5-VL-7B    & 12.5  & 137.9s  & 35.9\% \\
MedLVR           & 7.8   & 158.0s  & 45.2\% \\
\bottomrule
\end{tabular}
}
\end{table}

\subsection{Effect of the Latent Reasoning Budget}
\label{sec:latent_budget}

\begin{figure*}[tbp]
    \centering
    \includegraphics[width=\textwidth]{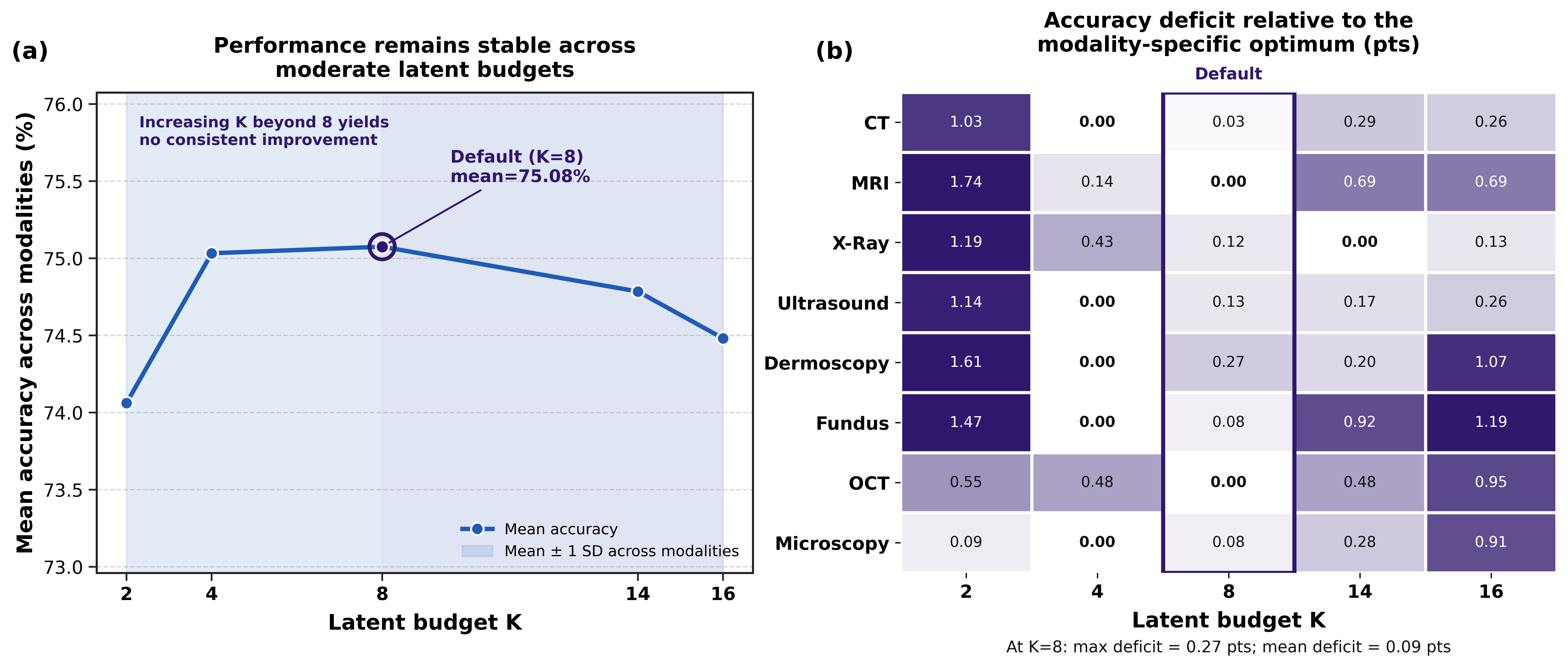}
    \caption{\textbf{Sensitivity to the latent reasoning budget.}
    \textbf{a}, Mean accuracy across eight imaging modalities under
    $K\in\{2,4,8,14,16\}$. Performance remains stable across moderate
    latent budgets, and increasing $K$ beyond 8 does not provide a
    consistent improvement. The shaded region denotes $\pm1$ standard
    deviation across modalities.
    \textbf{b}, Accuracy deficit relative to the modality-specific optimum
    for each latent budget. At the fixed default setting of $K=8$, the
    maximum deficit is 0.27 points and the mean deficit is 0.09 points,
    showing that a single shared budget retains near-optimal performance
    across modalities without modality-specific tuning.}
    \label{fig:latent_budget}
\end{figure*}

We next examine the sensitivity of \textsc{MedLVR} to the amount of latent
computation. The latent reasoning budget is varied across five settings,
$K\in\{2,4,8,14,16\}$. As shown in
Fig.~\ref{fig:latent_budget}(a), mean accuracy across the eight imaging
modalities improves substantially when the latent trajectory is extended
beyond a very small budget and remains stable across moderate settings.
In particular, $K=4$ and $K=8$ achieve nearly identical aggregate
performance, while increasing the budget beyond $K=8$ does not yield a
consistent improvement.

We further examine whether adopting a single fixed budget sacrifices
performance on individual imaging modalities. Figure~\ref{fig:latent_budget}(b)
reports the accuracy deficit of each setting relative to the best-performing
budget for each modality. At the default setting of $K=8$, the maximum deficit
across all eight modalities is only 0.27 points, while the mean deficit is
0.09 points. Thus, although the exact optimum varies slightly across modalities,
a single shared budget of $K=8$ retains performance close to
modality-specific tuning.

Based on this analysis, we use $K=8$ as the fixed latent reasoning budget for
all main experiments, including the modality-wise evaluation in
Table~\ref{tab:modality_comparison}. This choice avoids modality- or
benchmark-specific tuning while providing a stable balance between predictive
performance and additional latent computation. The $K=12$ rollout used in
Section~\ref{sec:mechanistic_validation} is reserved exclusively for
trajectory-resolved diagnostic analysis, where the trained checkpoints remain
unchanged and the extended rollout provides finer temporal resolution of the
latent process.

\subsection{Efficiency of Internal Visual Reasoning}
\label{sec:efficiency}

We finally examine the computational trade-off introduced by internal latent
reasoning. The comparison uses a 485-sample subset drawn from five medical VQA
benchmarks: 100 samples each from SLAKE, VQA-RAD, PMC-VQA, and MMMU
(Health \& Medicine), together with 85 samples from MedXpertQA.

As shown in Table~\ref{tab:efficiency_accuracy}, the compared methods allocate
reasoning computation in substantially different ways. MedGemma-1.5 generates
102.1 visible output tokens per sample and requires 1300.2 seconds, yet reaches
9.1\% overall accuracy on this evaluation subset. In contrast,
\textsc{MedLVR} achieves the highest accuracy of 45.2\% while generating only
7.8 visible output tokens per sample. The shorter textual output does not imply that \textsc{MedLVR} eliminates
additional reasoning computation. Its fixed latent trajectory introduces an
internal inference cost, increasing the total runtime from 137.9 seconds for
the Qwen2.5-VL-7B backbone to 158.0 seconds. This corresponds to a modest
runtime increase while improving accuracy from 35.9\% to 45.2\%.
\textsc{MedLVR} also remains substantially faster than MedGemma-1.5 and
requires a similar order of inference time to the other compact reasoning
systems in the comparison.

These results illustrate the computational role of latent visual reasoning.
Rather than expressing additional reasoning through a long visible textual
trajectory, \textsc{MedLVR} allocates a short, fixed amount of computation to
an internal latent process before answer generation. The resulting improvement
therefore comes with a measurable but limited runtime overhead, without
lengthening the visible response or requiring repeated external visual
operations.

\section{Discussion}

The preceding experiments establish that latent visual reasoning consistently improves medical VQA performance across imaging modalities and external benchmarks.
In this section we interpret these findings, discuss their clinical scope, acknowledge the limitations of the current study, and outline directions for future work.

\subsection{Interpretation of Findings}
\label{sec:interpretation}

The modality-level results in Table~\ref{tab:modality_comparison}
reveal two distinct patterns that should not be conflated:
the magnitude of improvement over the backbone, and the final
absolute accuracy achieved after training.

\paragraph{Distinguishing improvement magnitude from final accuracy.}
Ultrasound exhibits one of the largest absolute improvements,
increasing from 30.7\% to 51.8\% (+21.1 points), and OCT improves
from 61.2\% to 80.2\% (+19.0 points). Both gains are substantial
and comparable in magnitude to those observed on CT (+20.0 points)
and MRI (+23.6 points). The key difference is not in the size of
the improvement but in the final accuracy level reached:
ultrasound remains at 51.8\%, considerably below CT (80.4\%) and
MRI (82.0\%). These two dimensions, improvement magnitude and
final performance, reflect different aspects of the method and
should be interpreted separately.

\paragraph{Why final accuracy remains lower on ultrasound.}
One possible explanation is that ultrasound images present
inherently greater visual variability. Speckle noise,
operator-dependent probe positioning, and variable acoustic
shadowing can produce images with less spatially stable lesion
boundaries compared to CT or MRI under standard acquisition
protocols. However, we cannot isolate this effect from a
confounding factor: ultrasound accounts for only 1.14\% of the
Stage~1 grounding corpus, compared to 24.14\% for CT and 14.96\%
for MRI. The lower final accuracy on ultrasound may therefore
reflect the inherent visual complexity of the modality, the
limited availability of region-level supervision during training,
or most likely an interaction between the two. Disentangling
these factors would require controlled experiments with balanced
modality sampling, which we leave for future work.

\paragraph{CT and MRI.}
The high final accuracy on CT and MRI is consistent with
several favorable conditions, though the relative contribution
of each is difficult to isolate. These modalities benefit from
dense region-level supervision in Stage~1 and are well
represented in the Stage~2 OmniMedVQA training distribution.
Additionally, standard CT and MRI acquisitions may produce
images in which lesion boundaries and tissue contrast are
relatively well defined, creating a favorable setting for
ROI-based latent evidence formation. We note, however, that
imaging characteristics vary substantially across CT and MRI
protocols, sequences, and clinical sites, and the favorable
conditions described here should be understood as tendencies
rather than guarantees.

\paragraph{Broader pattern.}
The modality-level variation suggests that the final accuracy
achieved by \textsc{MedLVR} depends on at least two interacting
factors that the current experimental design cannot fully
disentangle: the degree to which visual evidence in a given
modality can be captured by spatially coherent patterns amenable
to ROI-based supervision, and the volume of region-supervised
examples available to shape the latent trajectory for that modality.
This interpretation is consistent with the mechanism analysis in
Section~\ref{sec:mechanistic_validation}, where ROI-specific evidence gaps are largest when
the latent trajectory can reliably distinguish the expert-defined
region from equally sized control regions.

\subsection{Clinical Scope}
\label{sec:clinical_scope}

MedLVR is designed for closed-ended or short-answer medical VQA
tasks in which the answer depends on localizable visual evidence
within a single image or a 2D slice from a volumetric study. The
framework has been evaluated exclusively on publicly available
academic benchmarks and has not been validated under real clinical
workflows, with prospective patient data, or in time-critical
point-of-care settings. Its current scope does not include
unrestricted clinical reasoning, volumetric image interpretation,
or autonomous clinical deployment.

\subsection{Limitations}
\label{sec:limitations}

\paragraph{Data supervision and coverage.}
The Stage~1 grounding corpus is constructed through a text-only
generation pipeline in which MedGemma receives the region label and
PrimeKG-retrieved context but does not observe the specific image.
This design reduces perceptual hallucination but anchors the
generated questions to category-typical visual attributes rather
than instance-specific observations; questions referencing precise
size, laterality, or spatial extent may not consistently match the
individual image. The expert quality audit covers 1{,}200 of the
505{,}554 automatically generated examples, providing a reasonably
precise estimate of the sampled corpus quality but leaving most
individual samples unreviewed. The corpus is intentionally
unbalanced to reflect natural data availability: CXR and CT account
for over 82\%, while dermoscopy, endoscopy, fundus photography, and
OCT each contribute less than 1\%, and microscopy is absent
entirely from Stage~1. Although Stage~2 reinforcement learning on
OmniMedVQA partially improves cross-modality balance, the latent
pathway receives substantially less region-level shaping for
low-frequency and absent modalities.

\paragraph{Generalizability of the evaluation.}
All experiments use a single backbone (Qwen2.5-VL-7B-Instruct),
leaving open whether the proposed training strategy transfers to
other model families, larger parameter counts, or different
vision--language integration mechanisms. Evaluation is conducted on
public academic benchmarks only; the method has not been tested with
data from external hospital systems or diverse patient populations,
where distribution shifts in imaging equipment and acquisition
protocols may affect performance. Although Section~3.5 provides a
systematic dataset-level overlap analysis, individual image-level
overlap between the grounding corpus and evaluation benchmarks such
as SLAKE and OmniMedVQA cannot be fully excluded. The quantitative
evidence from overlap-free benchmarks supports generalization, but
residual overlap risk remains. Additionally, comparisons with
several previously published methods rely on values reported in
their original papers because executable code or model weights were
unavailable; differences in prompting, preprocessing, and answer
parsing may affect direct comparability. Our principal conclusions
are based on controlled comparisons using the same backbone,
training data, test instances, and evaluation pipeline.

\paragraph{Spatial and computational design.}
Region-level supervision is provided through axis-aligned bounding
boxes derived from dataset-provided expert annotations, which are
inherently coarse approximations of the true diagnostic evidence.
For irregularly shaped lesions, diffuse pathological changes, or
findings spanning non-contiguous regions, the bounding box may
include non-diagnostic tissue or miss portions of the relevant
evidence; finer-grained supervision such as pixel-level masks could
provide more precise evidence-formation targets. MedLVR processes
2D images or individual slices and does not capture cross-slice
spatial relationships or 3D anatomical continuity that are often
essential for clinical interpretation of CT and MRI studies. The
framework uses a single fixed latent budget $K{=}8$ for all samples
regardless of question difficulty or image complexity; while
Section~5.8 shows this setting is near-optimal on average,
individual cases may benefit from more or fewer latent steps.

\paragraph{Interpretation of latent states.}
The mechanistic analyses in Section~5.4 demonstrate that latent
states become more aligned with expert-annotated regions and that
predictions are more sensitive to the removal of clinically relevant
image content. However, alignment in representation space should not
be equated with clinical interpretability. The latent trajectory
remains a high-dimensional continuous process without a
human-readable intermediate output, and the representational
similarity measures used to probe this process are mathematical
constructs that may not correspond to the clinical reasoning a human
expert would articulate. The current framework also does not produce
calibrated confidence scores or uncertainty estimates, limiting its
ability to identify unreliable predictions at inference time.

\subsection{Future Work}
\label{sec:future_work}

\paragraph{3D volumetric reasoning.}
Extending latent visual reasoning from 2D slices to full 3D volumes is a natural next step.
This would require designing latent trajectories that can aggregate evidence across multiple slices while maintaining computational feasibility, potentially through hierarchical latent structures that first reason within slices and then integrate cross-slice information.
Such an extension would address a fundamental limitation of the current framework and enable reasoning about volumetric findings such as tumor staging, organ segmentation, and multi-planar anatomical relationships.

\paragraph{Adaptive latent budget.}
Replacing the fixed latent budget with an adaptive mechanism that allocates computation based on input complexity could improve both efficiency and accuracy.
Possible approaches include learned halting criteria analogous to adaptive computation time, confidence-based early termination, or difficulty-aware budget allocation informed by the question and initial visual features.
An adaptive budget would allow simple cases to proceed with minimal latent computation while providing additional reasoning capacity for diagnostically challenging inputs.

\paragraph{Uncertainty estimation and calibrated abstention.}
Integrating uncertainty quantification into the latent reasoning process is critical for clinical applicability.
The stochastic latent policy used during Stage~2 training provides a natural starting point: the variance of latent trajectories sampled from the Gaussian policy could serve as a proxy for model uncertainty, and samples with high trajectory variance could be flagged for human review.
More formally, calibrated confidence scores and principled abstention mechanisms would enable the model to defer to human experts on cases where the latent evidence formation process does not converge.

\paragraph{Multi-lesion and multi-finding reasoning.}
The current ROI supervision assumes a single question-associated region per sample.
Extending the framework to handle multiple concurrent findings, such as co-occurring lesions in different anatomical regions or complex diagnostic patterns requiring the integration of evidence from several ROIs, would broaden clinical applicability.
This could involve multi-target latent evidence formation, where the trajectory is shaped by multiple ROI targets simultaneously, or sequential latent segments that address different findings in turn.

\paragraph{Prospective clinical evaluation.}
Ultimately, the clinical value of latent visual reasoning must be assessed through prospective studies in real clinical environments.
This includes evaluation on data from external hospital systems with diverse imaging equipment and patient populations, integration into clinical workflows to measure impact on diagnostic accuracy and efficiency, and studies of human--AI interaction to understand how clinicians engage with a model whose intermediate reasoning is not directly observable.
Such evaluation would provide evidence that extends beyond benchmark performance and addresses the translational gap between academic research and clinical deployment.

\section{Conclusion}
We presented \textsc{MedLVR}, a latent visual reasoning framework for medical VQA that introduces additional internal computation before answer generation. By combining ROI-supervised latent evidence formation with VLPO, \textsc{MedLVR} improves answer accuracy across multiple medical VQA benchmarks and increases the model's dependence on question-associated visual evidence. These results support latent visual reasoning as a complementary pathway for strengthening visual evidence use in medical VQA.


\bibliographystyle{cas-model2-names}
\bibliography{cas-refs}

\end{document}